\pdfoutput=1

\documentclass[11pt]{article}

\usepackage[preprint]{acl}

\usepackage{times}
\usepackage{latexsym}

\usepackage[T1]{fontenc}

\usepackage[utf8]{inputenc}

\usepackage{microtype}

\usepackage{inconsolata}

\usepackage{graphicx}
\usepackage{amsfonts}
\usepackage{amsmath} 
\usepackage{booktabs}
\usepackage[table]{xcolor}
\usepackage{float} 
\usepackage{array}
\newcolumntype{C}[1]{>{\centering\arraybackslash}m{#1}}
\usepackage{caption}
\title{\raisebox{-0.5em}{\includegraphics[height=1.8em]{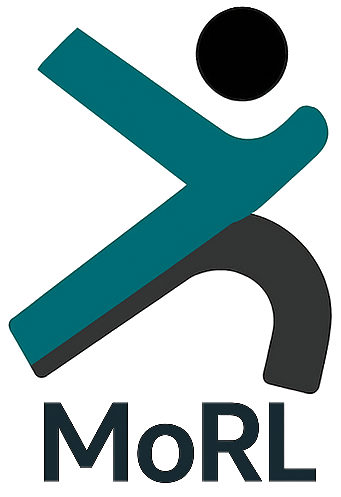}}~MoRL: Reinforced Reasoning for Unified Motion Understanding and Generation}

\author{
  \textbf{Hongpeng Wang$^{1*}$~~
  Zeyu Zhang$^{2*\dag}$~~
  Wenhao Li$^{3}$~~
  Hao Tang$^{2\ddag}$}\\
  $^1$The University of Sydney~~
  $^2$Peking University~~
  $^3$Nanyang Technological University\\
  \small$^{*}$Equal contribution. $^{\dag}$Project lead. $^{\ddag}$Corresponding author: bjdxtanghao@gmail.com.
}

\begin{document}
\maketitle
\begin{abstract}
Human motion understanding and generation are crucial for vision and robotics but remain limited in reasoning capability and test-time planning. We propose MoRL, a unified multimodal motion model trained with supervised fine-tuning and reinforcement learning with verifiable rewards. Our task-specific reward design combines semantic alignment and reasoning coherence for understanding with physical plausibility and text–motion consistency for generation, improving both logical reasoning and perceptual realism. To further enhance inference, we introduce Chain-of-Motion (CoM), a test-time reasoning method that enables step-by-step planning and reflection. We also construct two large-scale CoT datasets, MoUnd-CoT-140K and MoGen-CoT-140K, to align motion sequences with reasoning traces and action descriptions. Experiments on HumanML3D and KIT-ML show that MoRL achieves significant gains over state-of-the-art baselines.
Code: \url{https://github.com/AIGeeksGroup/MoRL}.
Website: \url{https://aigeeksgroup.github.io/MoRL}.

\end{abstract}

\section{Introduction}

Human motion understanding and generation are fundamental problems in computer vision and robotics. They enable a wide range of applications, from interactive character animation and robotics to game development and virtual reality. With the advent of large-scale motion capture datasets and expressive parametric human models such as SMPL \citep{loper2023smpl} and SMPL-X \citep{pavlakos2019expressive}, recent years have witnessed rapid progress in text-to-motion generation \citep{zhang2024motion,zhang2024infinimotion,zhang2025motion,zhang2024motionavatar,zhang2024kmm,zhang2025flashmo,wang2026safemo} and motion-language alignment \citep{zhang2023finemogen,guo2022generating}. Currently, the success of large language models (LLMs) has inspired multimodal extensions that integrate text, image, and 3D signals, pushing the frontier of motion language modeling toward more scalable and generalizable systems. Existing approaches have begun to explore this space. MotionGPT \citep{jiang2023motiongpt} considers motion as a foreign language to establish a unified action language framework. MotionRL \cite{liu2024motionrl}  introduces multi-reward optimization to better match human preferences. More recently, Motion-R1 \citep{ouyang2025motion} applies Chain-of-Thought reasoning and reinforcement learning to motion generation. 
\begin{figure}[!t]
  \centering
\includegraphics[width=0.49\textwidth]{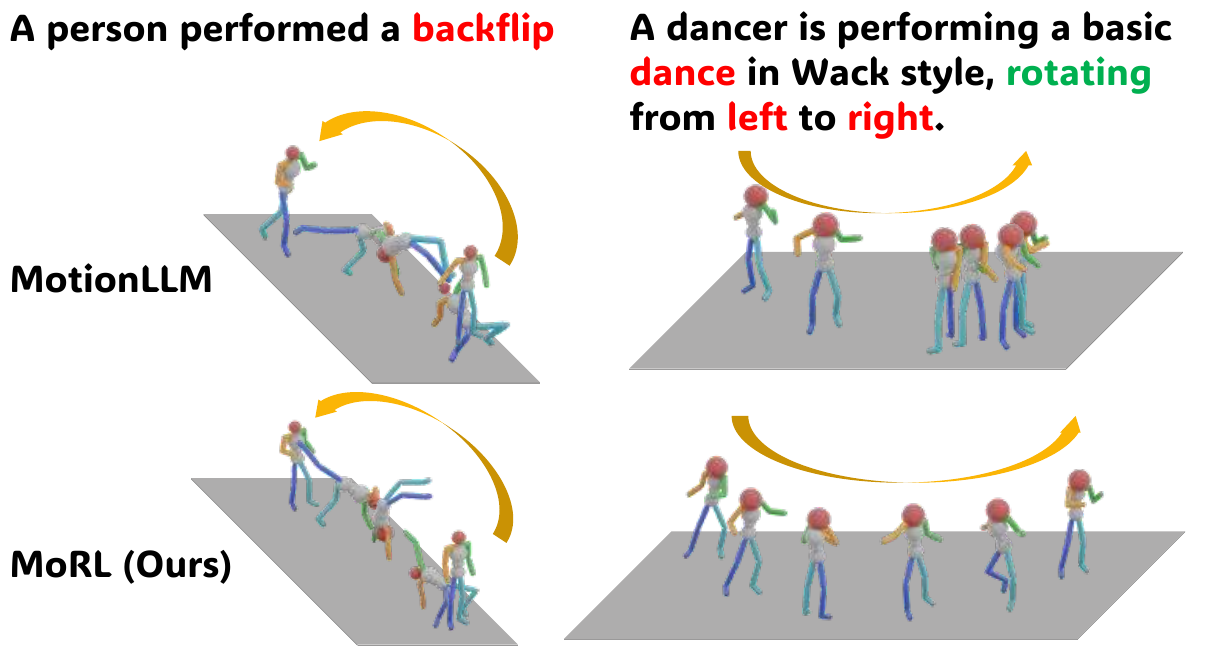} 
  \caption{Visualization comparisons with MotionLLM. In the backflip example, MotionLLM fails to maintain a coherent takeoff-rotation-landing trajectory, resulting in unstable body orientation, while MoRL completes a physically plausible flip. In the Wack-style dance, MotionLLM shows inconsistent rotation direction and fragmented poses, whereas MoRL preserves continuous left-to-right rotation and stylistic coherence.}
  \label{fig:vis1}
  \vspace{-0.5cm}
\end{figure}
Despite these advances, two major challenges remain. First, current models treat user queries as a whole, with limited reasoning capability. They struggle to parse prompts into fine-grained steps or to understand or generate detailed motions in a step-by-step manner. Second, at test time, most models simply decode outputs in a single pass. They lack explicit planning or reflection, and therefore cannot fully exploit the reasoning ability of large language models.   

To address the first challenge, we propose MoRL, a multimodal motion unified model that unifies motion understanding and generation under a reinforcement learning framework. MoRL is trained with a hierarchical post-training pipeline.  We then perform reinforcement learning with verifiable rewards (RLVR). Unlike prior works that rely primarily on generic similarity scores, our reward design is task-specific and dual-headed: for motion understanding, we introduce semantic alignment and a novel reasoning coherence reward that enforces logically consistent reasoning traces; for motion generation, we combine text–motion consistency with a physical plausibility reward that enforces biomechanical validity. This combination provides a simple yet innovative way to align model outputs with both semantic fidelity and human perceptual realism.

To address the second challenge,  and improve the test-time performance, we introduce Chain-of-Motion (CoM), a decoding strategy that explicitly incorporates step-by-step reasoning and reflection. CoM not only improves the robustness of reasoning-based motion understanding but also refines motion generation through iterative selection and correction. Moreover, the same principle guides the synthesis of our CoT datasets, ensuring consistency between training and inference.  Specifically, we construct two large-scale synthetic Chain-of-Thinking (CoT) datasets, MoUnd-CoT-140K and MoGen-CoT-140K, to align motion sequences with reasoning traces and concise action descriptions.
To further showcase the effectiveness, we conduct comprehensive experiments on HumanML3D \citep{guo2022generating} and KIT-ML \citep{plappert2016kit}. Results show that MoRL achieves significant gains over SOTA baselines.  

In summary, the main contributions are:  
\begin{itemize}
\item We propose MoRL, a unified multimodal motion model with task-specific rewards that improve motion understanding via semantic alignment and reasoning coherence, and motion generation via physical plausibility and text-motion consistency.
\item We introduce Chain-of-Motion, a test-time reasoning strategy, together with two large-scale CoT datasets, MoUnd-CoT-140K and MoGen-CoT-140K, to enhance motion understanding and generation through step-by-step reasoning and reflection.
\item Extensive experiments on HumanML3D and KIT-ML demonstrate that MoRL consistently outperforms state-of-the-art methods.

\end{itemize}

\section{Related Works}

\paragraph{Motion understanding and generation.}
Recent work on human motion understanding and generation has rapidly evolved from specialized sequence models to large language model (LLM)–based frameworks that unify perception, reasoning, and text–motion alignment.  
Early multimodal approaches such as MotionLLM \citep{chen2024motionllm}, ChatPose \citep{feng2024chatpose}, and ChatHuman \citep{lin2024chathuman} explored conversational or interactive motion generation, yet their evaluations largely focused on qualitative results without systematic motion-to-text benchmarking.  
UniMotion \citep{li2025unimotion} extended cross-modal modeling to a broader set of human activities, but it similarly omitted explicit motion-to-text evaluation, leaving the bidirectional mapping under-explored.  
LLM-driven pipelines such as MotionLLaMA \citep{ling2024motionllama} demonstrated impressive compositional motion synthesis but relied on private datasets, limiting reproducibility and large-scale comparison.  
Structured agent architectures like ACMo and CoMA \citep{sun2024coma} further highlighted the benefits of compositional reasoning and multi-modal interaction for controllable human-motion generation.  
Building on these foundations, a new wave of motion-generation systems integrates transformer backbones with LLM reasoning.  
Representative examples include MotionGPT \citep{zhang2024motiongpt,ribeiro2024motiongpt}, T2M-GPT \citep{wang2023t2m}, and ReMoGPT \citep{yu2025remogpt}, which leverage powerful language priors to improve both motion synthesis and natural-language controllability.  
Despite these advances, unified evaluation protocols that cover motion-to-text understanding, text-conditioned generation, and open-dataset benchmarking remain limited, motivating the need for methods that jointly address generation fidelity and cross-modal reasoning.

\begin{figure*}[t]
  \centering
  \includegraphics[width=1\textwidth]{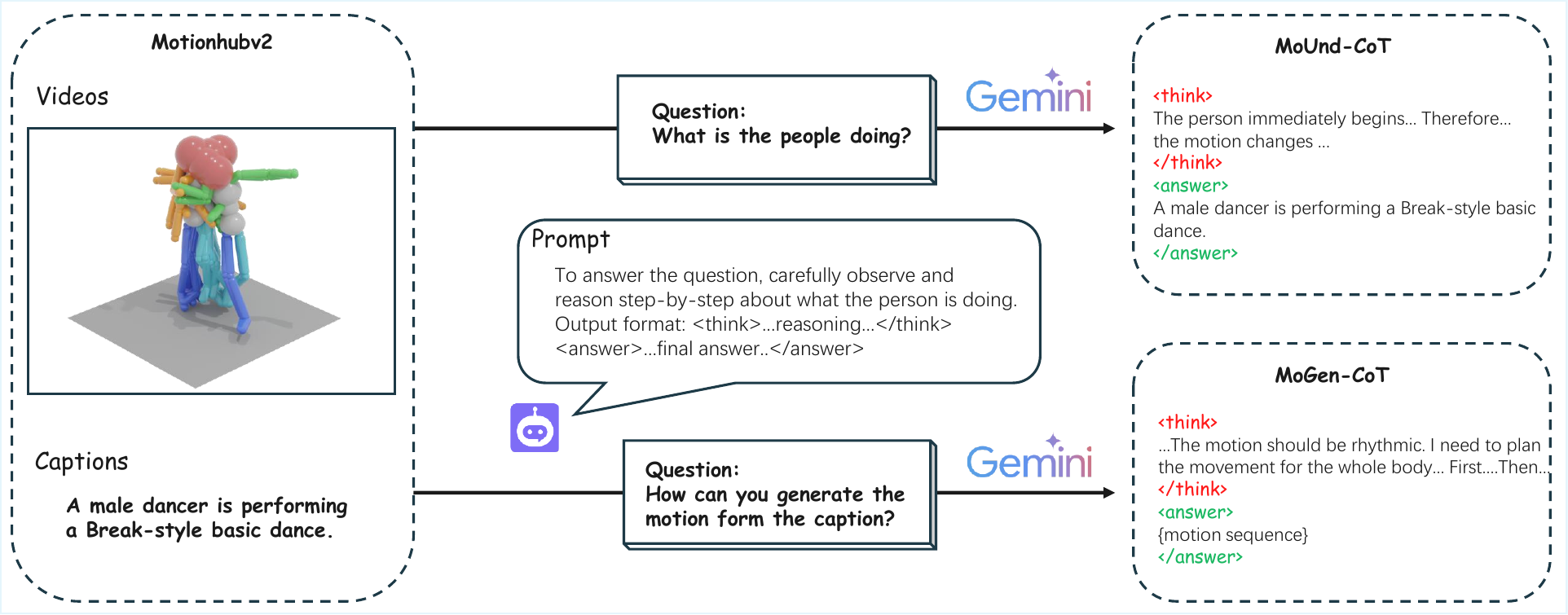} 
  \caption{Motion CoT data engine. Build based on MotionHubV2 dataset \citep{ling2024motionllama}, one branch (MoUnd-CoT) uses motion sequences and captions with Gemini to construct reasoning chains for understanding, while the other (MoGen-CoT) builds reasoning chains for generation.}
  \label{fig:cot_data_engine}
  \vspace{-0.5cm}
\end{figure*}

\paragraph{Large language model reasoning.} 

Many studies aim to enhance the reasoning capacity of Large Language Models (LLMs) to perform complex, multi-step problem-solving tasks by employing Chain-of-Thought (CoT) prompting \citep{wei2022chain,zhang2023multimodal,zhang2024improve,mitra2024compositional,hao2024training,yao2023tree,yuan2024advancing,luan2024textcot} and conducting supervised fine-tuning (SFT) with step-level supervision \citep{zhang2024llama,zhao2024marco,yao2024mulberry,thawakar2025llamav}. Recently, DeepSeek-R1 \citep{guo2025deepseek} successfully applied rule-based Reinforcement Learning (RL) \citep{shao2024deepseekmath} to induce the self-emergence of complex cognitive reasoning abilities in LLMs, demonstrating that even coarse, outcome-only rewards can effectively elicit strong reasoning behavior. Its success demonstrated that, with a carefully designed reward structure and policy optimization strategy, models can learn to generate long CoT reasoning without the need for intermediate supervision. Building on this paradigm, recent efforts such as Open-Reasoner-Zero \citep{hu2025open} and Kimi k1.5 \citep{team2025kimi} have adopted similar rule-based reinforcement learning pipelines to enhance reasoning in the text and image domains, respectively. However, despite these promising developments, little prior work has investigated extending this approach to the video domain. Bridging this gap remains both a significant challenge and a promising direction for advancing the capabilities of reasoning models.

\section{Data Synthesis}

\noindent\textbf{Data engine.}
The key to empowering MoRL with strong reasoning ability lies in large-scale, high-quality chain-of-thought (CoT) data. To address this gap, we design a data engine, as shown in Figure \ref{fig:cot_data_engine}, built on Gemini-2.5-pro \citep{comanici2025gemini}. It performs gap-based reasoning through question–answer pairs and captures the reasoning process. This aligns motion sequences with natural language reasoning chains and concise action captions. The sequences and captions are derived from the MotionHubV2 dataset \citep{ling2024motionllama}, which is constructed as a subset of multiple publicly available datasets and encompasses diverse motion scenarios such as dance, performance interaction, and various activities from daily life. The resulting dataset consists of two complementary branches: Motion Understanding and Motion Generation. Together, they form a unified CoT resource.

\noindent\textbf{MoUnd-CoT-140K.}
The motion understanding branch, denoted as \textit{MoUnd-CoT-140K}, is designed to map motion sequences into textual reasoning and descriptive outputs. Each data sample contains three components: (i) a motion sequence represented in the standard SMPL-X format, (ii) a reasoning chain enclosed in \texttt{<think>} tags, and (iii) a concise caption of the action enclosed in \texttt{<answer>} tags. To ensure compatibility with HumanML3D-style features, we convert SMPL-X joint sequences into humanml joint sequences and then extract motion features of dimension 263 per frame. This allows the dataset to be directly consumed by existing motion-language models. The resulting MoUnd-CoT-140K dataset provides high-quality CoT supervision for motion understanding tasks, especially in scenarios where the model must first interpret motion dynamics before generating a compact description. 

\begin{figure*}[t]
  \centering
  \includegraphics[width=1\textwidth]{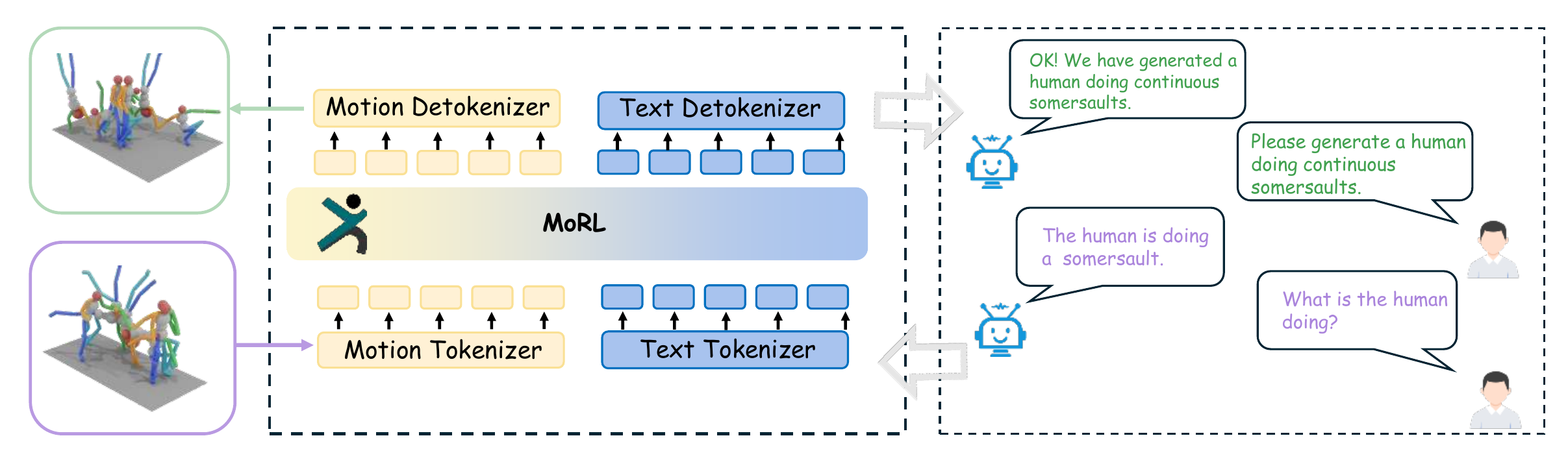} 
  \caption{Overview of MoRL. Our framework unifies motion understanding and generation under a reinforcement learning paradigm. Motion and text inputs are tokenized into a shared representation space. A hierarchical post-training pipeline first applies SFT on large-scale synthetic CoT datasets to align motion sequences with reasoning traces and concise descriptions, then employs reinforcement learning with verifiable rewards (RLVR) to refine outputs, enhancing semantic alignment, reasoning coherence, physical plausibility, and text–motion consistency. At inference, the Chain-of-Motion (CoM) decoding strategy enables step-by-step reasoning and reflection, improving both motion understanding and perceptually realistic motion generation.}
  \label{fig:overall}
  \vspace{-0.5cm}
\end{figure*}

\noindent\textbf{MoGen-CoT-140K.}
The motion generation branch, denoted as \textit{MoGen-CoT-140K}, complements MoUnd-CoT-140K by focusing on the inverse process: generating motion sequences from textual reasoning and descriptive inputs. Each sample contains (i) a natural language caption of the intended action, (ii) an associated reasoning chain in \texttt{<think>} tags, and (iii) the corresponding motion sequence stored in SMPL-X format contained between \texttt{<answer>} tags. For consistency, all sequences are normalized into the HumanML3D feature space. MoGen-CoT-140K thus enables motion-language models to learn not only to understand motion but also to generate realistic, semantically aligned motion sequences guided by reasoning signals. 

Together, MoUnd-CoT-140K and MoGen-CoT-140K form a balanced CoT-based motion-language corpus, enabling instruction tuning for both understanding and generation in a unified framework. 

\section{The Proposed Method}

\subsection{Overview}
As shown in Figure~\ref{fig:overall}, we propose MoRL, a multimodal motion foundation model that unifies human motion understanding and generation within a single framework. MoRL is built on a multimodal large language model (MLLM) initialized from Qwen3-4B-Instruct \cite{yang2025qwen3}, and augmented with dedicated text and motion tokenizers for cross-modal alignment. The framework comprises three key components: (1) a supervised fine-tuning (SFT) stage using a synthetic chain-of-thought (CoT) dataset for cold-start initialization; (2) task-specific reinforcement learning (RL) policies for motion understanding and motion generation, each optimized with tailored reward functions; and (3) a test-time reasoning strategy, COM, enhancing both understanding and generation through structured, step-by-step justification.

\subsection{Architecture}
MoRL adopts a unified multimodal LLM backbone equipped with two modality-specific tokenizers. The text tokenizer is inherited from the base language model, while the motion tokenizer discretizes continuous 3D human motion into compact motion tokens via a VQ-VAE style encoder–decoder. The multimodal fusion is achieved through shared transformer layers, enabling cross-attention between textual and motion representations. This design follows the paradigm of motion-language alignment in DeepSeek but extends it to bidirectional tasks, including text-to-motion generation and motion-to-text understanding.

\noindent\textbf{Text tokenizer.}  
We employ the native tokenizer of the LLM to map natural language into subword tokens. This preserves the rich linguistic knowledge of the base LLM while ensuring compatibility with motion-related vocabulary introduced during supervised fine-tuning. The text tokens serve as both queries (in understanding tasks) and conditioning signals (in generation tasks).

\noindent\textbf{Motion tokenizer.}  
To bridge the gap between continuous human motion and the discrete token space of the LLM, we adopt a VQ-VAE style motion tokenizer. Given an input motion sequence 
$m_{1:T} \in \mathbb{R}^{T \times D}$,  
where $T$ is the number of frames and $D$ is the dimensionality of each frame, the encoder $E$ compresses the sequence into latent vectors $z_{1:(T/l)} \in \mathbb{R}^{(T/l)\times d}$ with downsampling factor $l$ and latent dimension $d$. Each latent $z_i$ is then quantized against a learnable codebook $\mathcal{C}=\{c_n\}_{n=1}^N$:
\begin{equation}
\hat{z}_i = \arg\min_{c_n \in \mathcal{C}} \|z_i - c_n\|_2^2.
\end{equation}

The quantized sequence $\hat{z}_{1:(T/l)}$ is decoded back to reconstruct the original motion $\hat{m}_{1:T} = D(\hat{z}_{1:(T/l)})$. Training follows the composite VQ-VAE loss:
\begin{equation}
\mathcal{L}_{vq} = \mathcal{L}_{reconstruct} + \mathcal{L}_{commit} + \mathcal{L}_{embed},
\end{equation}
where $\mathcal{L}_{reconstruct}$ is a smoothed L1 loss with velocity regularization, $\mathcal{L}_{commit}$ enforces codebook utilization, and $\mathcal{L}_{embed}$ stabilizes latent representations. This discrete motion representation not only reduces sequence length but also aligns seamlessly with the autoregressive generation paradigm of LLMs.

\subsection{Cold Start Stage}
Recent work such as DeepSeek-R1 \citep{guo2025deepseek} demonstrated that reinforcement learning alone can sometimes induce CoT reasoning. Motivated by this, we first explored training our motion–language model directly with RL signals. In practice, however, this strategy was highly unstable: the model rarely produced well-formed reasoning traces and even generated answers that deviated from the intended semantics. To stabilize training, we introduce a cold-start phase based on supervised fine-tuning. Specifically, we use our synthetic datasets MoUnd-CoT-140K and MoGen-CoT-140K, which couple motion sequences with reasoning steps (\texttt{<think>}) and concise descriptions (\texttt{<answer>}). Supervised finetuning on these data forces the model to follow the required output format, stabilizing its outputs and ensuring semantic consistency between inference and final answers. This initialization greatly reduces collapse during RL and establishes a reliable starting point for policy optimization.  

\subsection{Reinforcement Learning}
After cold-start training, we further align the model outputs with task objectives through reinforcement learning. We adopt a group-based policy optimization strategy similar to GRPO, where multiple candidate outputs are sampled per prompt, scored with reward functions, normalized within the group, and used to compute policy gradients with a KL regularization term to a frozen reference model.  

\noindent\textbf{Motion understanding.}
For motion understanding, the model must output a reasoning trace \(\hat{r}\) and a caption \(\hat{a}\) given a motion sequence \(m\). We define two rewards:  

\textit{Semantic Alignment Reward}.  
We measure the semantic similarity between \(\hat{a}\) and the reference caption \(a\) using a pretrained text encoder \(E_{\text{text}}\):  
\begin{equation}
R_{\text{sem}} = \cos\!\left( E_{\text{text}}(\hat{a}), E_{\text{text}}(a) \right).
\end{equation}  

\textit{Reasoning Coherence Reward}. 
We encourage the reasoning trace to logically support the answer using an NLI model \(f_{\text{NLI}}\):  
\begin{equation}
R_{\text{coh}} = f_{\text{NLI}}(\hat{r}, \hat{a}),
\end{equation}  
where we implement $f_{\text{NLI}}(\cdot)$ using a frozen DeBERTa-v3-large MNLI model and take its entailment probability as the coherence score. $f_{\text{NLI}}(\cdot)$ outputs an entailment confidence score.  

\noindent\textbf{Motion generation.}
For motion generation, the model produces a motion sequence \(\hat{m}\) from a text prompt \(t\). We use two rewards:  

\textit{Physical Plausibility Reward}.  
We penalize implausible motion dynamics:  
\begin{equation}
R_{\text{phys}} = - \lambda_1 \cdot L_{\text{joint}}(\hat{m}) 
- \lambda_2 \cdot L_{\text{vel}}(\hat{m}),
\end{equation}  
where $L_{\text{joint}}(\cdot)$ measures joint-angle violations and $L_{\text{vel}}(\cdot)$ penalizes abrupt velocity changes, and $\lambda_{1} = 0.8$ for joint-limit violation and $\lambda_{2} = 0.2$.  

\textit{Text–Motion Consistency Reward}.  
We enforce semantic alignment between generated motion and the input text, using cross-modal encoders \(E_{\text{text}}, E_{\text{motion}}\):  
\begin{equation}
R_{\text{align}} = \cos\!\left( E_{\text{text}}(t), E_{\text{motion}}(\hat{m}) \right).
\end{equation}

\subsection{Chain-of-Motion}

Most motion-language models decode outputs in a single pass, often resulting in shallow semantic reasoning for understanding and temporal inconsistency for generation. We propose Chain-of-Motion (CoM), a test-time reasoning strategy that introduces explicit step-by-step planning and reflection.

Given an input prompt or motion sequence, the model first generates an intermediate natural-language reasoning trace, analogous to Chain-of-Thought. For motion understanding, this trace explains causal and temporal structure to support the final caption; for motion generation, it outlines a sequence of action primitives before decoding motion tokens, guiding fine-grained dynamics.

CoM further samples multiple candidate reasoning-motion pairs and evaluates them using task-specific rewards (reasoning-answer coherence for understanding, and semantic alignment with physical plausibility for generation). Low-quality candidates are discarded, while high-quality ones are refined through iterative reflection, reducing semantic drift and physically implausible motions.

Finally, CoM is consistent with training: our MoUnd-CoT-140K and MoGen-CoT-140K datasets include explicit reasoning traces, enabling CoM to naturally extend the SFT and RL stages at inference time.

\section{Experiments}

\subsection{Experiments Settings}
\noindent\textbf{Datasets.}
We evaluate MoRL on two widely used motion–language benchmarks: HumanML3D \citep{guo2022generating} and KIT-ML \citep{plappert2016kit}.
HumanML3D contains 14.6K motion clips with 44.9K text annotations, covering diverse everyday actions, while KIT-ML includes 3.9K motions paired with 6.3K linguistically varied descriptions.
Following prior work, motions are represented using SMPL-based joint features (263 for HumanML3D and 251 for KIT-ML), with temporal normalization and left–right mirroring applied.
Both datasets are split into training, validation, and test sets with a ratio of 0.8/0.15/0.05.

\noindent\textbf{Metrics.} 
For motion understanding, we adopt standard linguistic similarity metrics. BLEU@1 and BLEU@4 measure unigram and 4-gram precision, capturing lexical overlap. ROUGE-L evaluates longest common subsequence, reflecting recall-oriented alignment. CIDEr computes TF-IDF weighted $n$-gram consensus across references, rewarding semantic coverage. BERTScore uses contextual embeddings to assess semantic similarity beyond surface overlap.

For motion generation, we follow established benchmarks. RPrecision (Top1/Top2/Top3) measures cross-modal retrieval accuracy between motion and text. FID evaluates the distributional gap between generated and real motions, where lower is better. MM Dist measures motion–text embedding distance in a shared space. Diversity quantifies variation across generated motions from different prompts. MModality evaluates the ability to produce distinct yet semantically consistent motions under the same text prompt.

\begin{table*}[t]
\centering
\renewcommand{\arraystretch}{0.9}
\resizebox{\textwidth}{!}{%
\begin{tabular}{lccccccc|ccccc}
\toprule
\textbf{Method} 
& \multicolumn{7}{c}{\textbf{Motion Generation}}
& \multicolumn{5}{c}{\textbf{Motion Understanding}} \\
\cmidrule(lr){2-8} \cmidrule(lr){9-13}
& R@1$\uparrow$ & R@2$\uparrow$ & R@3$\uparrow$
& FID$\downarrow$ & MM Dist$\downarrow$
& Div$\rightarrow$ & MM$\uparrow$
& BLEU@1$\uparrow$ & BLEU@4$\uparrow$
& ROUGE-L$\uparrow$ & CIDEr$\uparrow$
& BERTScore$\uparrow$ \\
\midrule

\multicolumn{13}{c}{\textbf{HumanML3D} \citep{guo2022generating}} \\
\midrule

GT / Real Motions
& 0.511 & 0.703 & 0.797 & 0.002 & 2.974 & 9.503 & -
& - & - & - & - & - \\

SeqGAN \citep{goutsu2021linguistic}
& - & - & - & - & - & - & -
& 47.80 & 13.50 & 39.20 & 50.20 & 23.40 \\

RAEs \citep{yamada2018paired}
& - & - & - & - & - & - & -
& 33.30 & 10.20 & 37.50 & 22.10 & 10.70 \\

Seq2Seq(Att) \citep{plappert2018learning}
& - & - & - & - & - & - & -
& 51.80 & 17.90 & 46.40 & 58.40 & 29.10 \\

T2M \citep{guo2022generating}
& 0.457 & 0.639 & 0.740 & 1.067 & 3.340 & 9.188 & 2.090
& - & - & - & - & - \\

T2M-GPT \citep{zhang2023t2mgpt}
& 0.491 & 0.680 & 0.775 & 0.116 & 3.118 & 9.761 & 1.856
& - & - & - & - & - \\

FineMoGen \citep{zhang2023finemogen}
& 0.504 & 0.690 & 0.784 & 0.151 & 2.998 & 9.263 & 2.696
& - & - & - & - & - \\

MoGenTS \citep{yuan2024mogents}
& 0.529 & 0.719 & 0.812 & 0.033 & 2.867 & 9.570 & -
& - & - & - & - & - \\

Language2Pose \citep{ahuja2019language2pose}
& 0.246 & 0.387 & 0.486
& 11.02 & 5.296 & 7.676 & -
& - & - & - & - & - \\

ReMoDiffuse \citep{zhang2023remodiffuse}
& 0.510 & 0.698 & 0.795 & 0.103 & 2.974 & 9.018 & 1.795
& - & - & - & - & - \\

ReMoGPT \citep{yu2024remogpt}
& 0.501 & 0.688 & 0.792 & 0.205 & 2.929 & 9.763 & 2.816
& - & - & - & - & - \\

RMD \citep{liao2024rmd}
& 0.524 & 0.715 & 0.811 & 0.111 & 2.879 & 9.527 & 2.604
& - & - & - & - & - \\

MoRAG-Diffuse \citep{kalakonda2024morag}
& 0.511 & 0.699 & 0.792 & 0.270 & 2.950 & 9.536 & 2.773
& - & - & - & - & - \\

Lyu et al. \citep{lyu2025towards}
& - & - & - & - & - & - & -
& 49.70 & 13.62 & 39.20 & 53.10 & 33.10 \\

MDM \citep{tevet2023mdm}
& - & - & 0.611 & 0.544 & 5.566 & 9.559 & 2.799
& - & - & - & - & - \\

MotionDiffuse \citep{zhang2024motiondiffuse}
& 0.491 & 0.681 & 0.782 & 0.630 & 3.113 & 9.410 & 1.553
& - & - & - & - & - \\

Motion2Language \citep{radouane2024motion2language} & - & - & - & - & - & - & - & 67.00 & 23.40 & 53.80 & 53.70 & 37.20 \\

M2T-Interpretable \citep{radouane2023guided}
& - & - & - & - & - & - & -
& 69.90 & 25.00 & 55.30 & 61.60 & 40.30 \\

Text2Gesture \citep{bhattacharya2021text2gestures}
& 0.165 & 0.267 & 0.345
& 7.664 & 6.030 & 6.409 & -
& - & - & - & - & - \\

MoMask \citep{guo2024momask}
& 0.521 & 0.713 & 0.807 & 0.045 & 2.958 & - & 1.241
& - & - & - & - & - \\

ReMoMask \citep{li2025remomask}
& 0.531 & 0.722 & 0.813 & 0.099 & 2.865 & 9.535 & 2.823
& - & - & - & - & - \\

\rowcolor{yellow!30}TM2T\citep{guo2022tm2t}
& 0.424 & 0.618 & 0.729 & 1.501 & 3.467 & 8.589 & 2.424
& 61.70 & 22.30 & 49.20 & 72.50 & 37.80 \\

\rowcolor{yellow!30}TM2T$^*$ \citep{guo2022tm2t}
& 0.424 & 0.618 & 0.729 & 1.501 & 3.467 & 8.589 & 2.424
& 48.90 & 8.270 & 38.10 & 15.80 & 32.20 \\

\rowcolor{yellow!30}AvatarGPT \citep{zhou2024avatargpt}
& 0.510 & 0.702 & 0.796 & 0.168 & - & 9.624 & -
& 49.28 & 12.70 & 40.44 & 32.65 & \textbf{53.58} \\

\rowcolor{yellow!30}MotionGPT \citep{jiang2023motiongpt}
& 0.492 & 0.681 & 0.733 & 0.232 & 3.096 & 9.528 & 2.008
& 48.20 & 12.47 & 37.40 & 29.20 & 32.40 \\

\rowcolor{yellow!30}MotionGPT-2 \citep{wang2024motiongpt}
& 0.496 & 0.691 & 0.782 & \textbf{0.191} & 3.080 & \textbf{9.860} & 2.137
& 48.70 & 13.80 & 37.60 & 29.80 & 32.60 \\

\rowcolor{yellow!30}MotionChain \citep{jiang2024motionchain}
& 0.504 & - & 0.790 & 0.248 & 3.033 & 9.470 & -
& 48.10 & 12.56 & 33.90 & 33.70 & 36.90 \\

\rowcolor{yellow!30}Motion Agent \citep{wu2024motion}
& 0.515 & - & 0.801 & 0.230 & 2.967 & 9.908 & -
& 54.53 & 17.65 & 48.70 & 33.74 & 42.63 \\

\rowcolor{yellow!30}LaMP \citep{li2025lamp}
& 0.557 & 0.751 & 0.843 & 0.032 & 2.759 & 9.571 & -
& 47.80 & 13.04 & 37.10 & 28.90 & 32.70 \\

\midrule
\rowcolor{yellow!60}\textbf{MoRL (Ours)}
& \textbf{0.527} & \textbf{0.711} & \textbf{0.821}
& 0.203 & \textbf{2.790} & 9.701 & \textbf{2.702}
& \textbf{56.99} & \textbf{20.54} & \textbf{51.83}
& \textbf{35.80} & 46.80 \\

\midrule

\multicolumn{13}{c}{\textbf{KIT-ML} \citep{plappert2016kit}} \\
\midrule

Real Motions
& 0.424 & 0.649 & 0.779 & 0.031 & 2.788 & 11.08 & -
& - & - & - & - & - \\

SeqGAN \citep{goutsu2021linguistic}
& - & - & - & - & - & - & -
& 3.12 & 5.20 & 32.40 & 29.50 & 2.20 \\

RAEs \citep{yamada2018paired}
& - & - & - & - & - & - & -
& 30.60 & 0.10 & 25.70 & 8.00 & 0.40 \\

Seq2Seq(Att) \citep{plappert2018learning}
& - & - & - & - & - & - & -
& 34.30 & 9.30 & 36.30 & 37.30 & 5.30 \\

T2M \citep{guo2022generating}
& 0.370 & 0.569 & 0.693 & 2.770 & 3.401 & 10.91 & 1.482
& - & - & - & - & - \\

T2M-GPT \citep{zhang2023t2mgpt}
& 0.416 & 0.627 & 0.745 & 0.514 & 3.007 & 10.92 & 1.570
& - & - & - & - & - \\

MoGenTS \citep{yuan2024mogents}
& 0.445 & 0.671 & 0.797 & 0.143 & 2.711 & 10.918 & -
& - & - & - & - & - \\

ReMoDiffuse \citep{zhang2023remodiffuse}
& 0.427 & 0.641 & 0.765 & 0.155 & 2.814 & 10.80 & 1.239
& - & - & - & - & - \\

Language2Pose \citep{ahuja2019language2pose}
& 0.221 & 0.373 & 0.483
& 6.545 & 5.147 & 9.073 & -
& - & - & - & - & - \\

Lyu et al. \citep{lyu2025towards}
& - & - & - & - & - & - & -
& 43.40 & 8.90 & 35.20 & 65.30 & 31.20 \\

MDM \citep{tevet2023mdm}
& - & - & 0.396 & 0.497 & 9.191 & 10.85 & 1.907
& - & - & - & - & - \\

MotionDiffuse \citep{zhang2024motiondiffuse}
& 0.417 & 0.621 & 0.739 & 1.954 & 2.958 & 11.10 & 0.730
& - & - & - & - & - \\

Motion2Language \citep{radouane2024motion2language} & - & - & - & - & - & - & - & 56.80 & 25.40 & 58.80 & 125.7 & 42.10 \\

M2T-Interpretable \citep{radouane2023guided}
& - & - & - & - & - & - & -
& 58.40 & 24.40 & 58.30 & 112.1 & 41.20 \\

Text2Gesture \citep{bhattacharya2021text2gestures}
& 0.156 & 0.255 & 0.338
& 12.12 & 6.964 & 9.334 & -
& - & - & - & - & - \\

MoMask \citep{guo2024momask}
& 0.433 & 0.656 & 0.781 & 0.204 & 2.779 & - & 1.131
& - & - & - & - & - \\

ReMoMask \citep{li2025remomask}
& 0.453 & 0.682 & 0.805 & 0.138 & 2.682 & 10.83 & 2.017
& - & - & - & - & - \\

\rowcolor{yellow!30}TM2T \citep{guo2022tm2t}
& 0.280 & 0.463 & 0.587 & 3.599 & 4.591 & 9.473 & \textbf{3.292}
& 46.70 & 18.40 & 44.20 & \textbf{79.50} & 23.00 \\

\rowcolor{yellow!30}TM2T$^*$  \citep{guo2022tm2t}
& 0.280 & 0.463 & 0.587 & 3.599 & 4.591 & 9.473 & \textbf{3.292}
& 35.10 & 6.200 & 28.70 & 28.90 & 30.40 \\

\rowcolor{yellow!30}MotionGPT \citep{jiang2023motiongpt}
& 0.366 & 0.558 & 0.680 & 0.510 & 3.527 & 10.350 & 2.328
& - & - & - & - & - \\

\rowcolor{yellow!30}MotionGPT-2 \citep{wang2024motiongpt}
& 0.427 & 0.627 & 0.764 & 0.614 & 3.164 & \textbf{11.256} & 2.357
& - & - & - & - & - \\

\rowcolor{yellow!30}LaMP \citep{li2025lamp}
& 0.479 & 0.691 & 0.826 & 0.141 & 2.704 & 10.929 & -
& - & - & - & - & - \\

\midrule
\rowcolor{yellow!60}\textbf{MoRL (Ours)}
& \textbf{0.439} & \textbf{0.661} & \textbf{0.793}
& \textbf{0.204} & \textbf{2.777} & 10.882 & 1.991
& \textbf{52.11} & \textbf{19.31} & \textbf{49.96}
& 34.04 & \textbf{33.66} \\

\bottomrule
\end{tabular}
}
\caption{Comparison of motion generation and motion understanding on HumanML3D and KIT-ML. Highlights indicate the unified model, bold  represent the best results within the unified model. Results marked with $^*$ are reproduced by MotionGPT~\citep{jiang2023motiongpt} and Lyu et al.~\citep{lyu2025towards}, and are computed with unprocessed ground truth texts for linguistic metrics.}
\label{tab:unified_all}
\vspace{-0.5cm}
\end{table*}

\noindent\textbf{Implementation details.}
Our backbone is initialized from Qwen3-4B-Instruct \cite{yang2025qwen3}, a compact yet capable language model. Motion sequences are first converted into frame-level features using the HumanML3D feature extractor, and then discretized by a VQ-VAE motion tokenizer. In practice, our motion tokenizer uses a codebook of $N=512$ entries and a latent dimension of 128.  The text is encoded with the Qwen tokenizer. To adapt the model efficiently, we insert LoRA adapters into the attention and feed-forward layers with rank $r=16$ and dropout 0.1.  

Training proceeds in two stages. In the SFT stage, we fine-tune on our synthetic CoT datasets (MoUnd-CoT-140K and MoGen-CoT-140K) with AdamW optimizer, learning rate $1\times10^{-5}$, batch size 64, and weight decay $0.01$ for 5 epochs. In the RL stage, we adopt group-based reinforcement learning with group size 8. Candidate outputs are scored with our reward functions, normalized within each group, and optimized using a KL-regularized objective toward a frozen SFT reference. The RL learning rate is $5\times10^{-6}$, and training is run for 3 epochs.

All models are trained in PyTorch on four NVIDIA A100 GPUs. During inference, we apply the Chain-of-Motion decoding strategy with $K=8$ candidates and $T=2$ refinement iterations, which adds only a modest runtime overhead while consistently improving output quality.

\subsection{Main Results}

\noindent\textbf{Motion understanding.}
Table~\ref{tab:unified_all} reports results on HumanML3D and KIT-ML understanding benchmarks. MoRL achieves consistent improvements across all linguistic metrics, outperforming both traditional sequence models (e.g., Seq2Seq(Att)~\citep{plappert2018learning}) and recent LLM-based methods such as MotionGPT~\citep{jiang2023motiongpt} and Motion Agent~\citep{wu2024motion}.
On HumanML3D, MoRL improves BLEU@1 and BLEU@4 by a clear margin over Motion Agent, while yielding higher ROUGE-L and BERTScore, indicating better semantic fidelity and more fluent language generation. Notably, MoRL reaches a CIDEr score of 35.8, substantially higher than Motion Agent (33.74), showing stronger consensus with human-annotated references.
On KIT-ML, MoRL also achieves the best balance between precision-oriented (BLEU) and semantic-oriented metrics (BERTScore, ROUGE-L), demonstrating that our dual reward design generalizes well across datasets. These gains primarily come from the semantic alignment and reasoning-coherence rewards, which ensure that generated descriptions are both logically consistent and well-grounded in motion semantics.

Notably, under comparisons among unified model approaches, our method achieves comprehensive superiority across most metrics, and even when compared to separate models, it attains comparable or superior performance on certain methods and metrics.

\noindent\textbf{Motion generation.}
We further evaluate MoRL on text-to-motion generation (Table~\ref{tab:unified_all}). On HumanML3D, MoRL consistently improves R-Precision across Top-1/2/3 over strong baselines such as ReMoGPT~\citep{yu2025remogpt} and MoRAG-Diffuse~\citep{kalakonda2024morag}, highlighting its superior text–motion alignment. Although FID is slightly higher than the best-performing diffusion-based models, MoRL achieves the lowest multimodal distance, suggesting closer alignment to reference motions in feature space. Moreover, MoRL delivers competitive diversity and strong multimodality, showing that our physical plausibility and text–motion consistency rewards encourage both realism and variety in generated motions.
On KIT-ML, MoRL achieves comparable performance to state-of-the-art diffusion models, with balanced R-Precision and FID values. While not always the absolute best in each metric, MoRL provides robust overall performance across fidelity, diversity, and alignment. Importantly, the introduction of Chain-of-Motion at test time further stabilizes inference, reducing error propagation and producing smoother, more natural motion trajectories.

Similarly, our method outperforms most unified models and shows notable advantages even compared to some separate models.

\subsection{Qualitative Analysis and Visualization}
Figure \ref{fig:vis1} presents gneneration-qualitative comparisons between MoRL and MotionLLM on two representative prompts.
For the simple caption describing a backflip (left), MotionLLM produces an incorrect global displacement: the preparatory bending phase drifts forward relative to the standing position, and the motion ends abruptly after the flip, resulting in an unnatural transition. In contrast, MoRL generates a complete and temporally coherent backflip, including a correct takeoff location, a smooth mid-air rotation, a stable landing, and a natural recovery sequence. The improved fidelity demonstrates MoRL’s stronger physical reasoning and its ability to handle high-momentum, highly dynamic motions. For the more complex caption describing a Wack-style dance (right), MotionLLM fails to maintain a consistent left-to-right rotational pattern and produces fragmented upper-body movements. MoRL outputs smoother, directionally consistent, and stylistically richer motions, accurately reflecting both the intended dance style and the global rotation described in the text. The incorporation of CoM-based reasoning further enhances motion naturalness and semantic grounding. These results indicate that our semantic alignment reward and CoM inference together improve long-horizon motion planning and text–motion correspondence.

\begin{table}[t]
\centering
\setlength{\tabcolsep}{3.2pt}
\small
\begin{tabular}{lccc cc}
\toprule
\textbf{Variant} 
& \multicolumn{3}{c}{\textbf{Understanding}}
& \multicolumn{2}{c}{\textbf{Generation}} \\
\cmidrule(lr){2-4} \cmidrule(lr){5-6}
& BERT$\uparrow$ & CIDEr$\uparrow$ & R-L$\uparrow$
& R@1$\uparrow$ & FID$\downarrow$ \\
\midrule
SFT only        & 42.65 & 33.88 & 48.78 & 0.420 & 0.212 \\
\midrule
w/o $R_{\text{sem}}$   & 44.10 & 34.05 & 50.01 & 0.488 & 0.209 \\
w/o $R_{\text{coh}}$   & 44.32 & 35.12 & 49.10 & 0.512 & 0.206 \\
w/o $R_{\text{phys}}$  & 46.18 & 35.50 & 51.19 & 0.518 & 0.285 \\
w/o $R_{\text{align}}$ & 45.00 & 34.62 & 50.48 & 0.492 & 0.225 \\
w/o CoM                & 45.48 & 34.98 & 50.78 & 0.505 & 0.220 \\
\midrule
\textbf{Full MoRL} 
& \textbf{46.80} & \textbf{35.80} & \textbf{51.83}
& \textbf{0.527} & \textbf{0.203} \\
\bottomrule
\end{tabular}
\caption{Ablation study of MoRL on HumanML3D.}
\label{table:ablation}
\vspace{-0.5cm}
\end{table}

\subsection{Ablation Study}
\noindent\textbf{Ablation on Model Components.}
We conduct ablation studies on HumanML3D to evaluate the contribution of each component in MoRL (Table~\ref{table:ablation}). Starting from the SFT-only baseline, which yields the weakest performance for both understanding and generation, progressively adding RLVR rewards and CoM consistently improves results.

Removing the semantic alignment reward ($R_{\text{sem}}$) notably degrades BERTScore and CIDEr, highlighting its role in grounding textual semantics. Excluding the reasoning coherence reward ($R_{\text{coh}}$) mainly affects ROUGE-L and CIDEr, confirming its importance for logical and temporal consistency. Dropping the physical plausibility reward ($R_{\text{phys}}$) preserves language metrics but significantly worsens FID, indicating its necessity for realistic motion synthesis. Removing the text–motion consistency reward ($R_{\text{align}}$) causes a substantial drop in R-Precision, revealing its role in cross-modal alignment. Finally, excluding CoM leads to moderate performance degradation across metrics, demonstrating its contribution to test-time reasoning.

\noindent\textbf{Comparison of Rewards.}
To compare the impact of the different reward designs, we first construct a Complex Motion Subset (CMS) from the HumanML3D dataset to evaluate motion generation under long temporal horizons and compositional semantic constraints. Specifically, we select samples from the original test set that satisfy the following criteria: (1) the textual description contains at least three action verbs (e.g., walk, turn, sit, raise); (2) the description includes explicit temporal connectors (e.g., then, after, finally, while), indicating clear ordering dependencies between actions; and (3) the text length is no fewer than 20 tokens. Samples in this subset typically correspond to multi-stage motions with long temporal spans, posing higher demands on global semantic modeling and long-range consistency.

We keep the model architecture, data, and optimization fixed and vary only the generation reward for a controlled comparison.
Table~\ref{tab:cms_reward} reports results on CMS.

\begin{table}[t]
\centering
\resizebox{\columnwidth}{!}{%
\begin{tabular}{lcccccc}
\toprule
\textbf{Generation Reward} 
& R@1$\uparrow$ & R@2$\uparrow$ & R@3$\uparrow$
& FID$\downarrow$ & MM Dist$\downarrow$ & MM$\uparrow$ \\
\midrule
Motion-R1 Reward \citep{ouyang2025motion} 
& 0.472 & 0.651 & 0.742 
& 0.185 & 3.021 & 2.31 \\

MotionRL Reward \citep{liu2024motionrl} 
& \textbf{0.491} & 0.676 & 0.768 
& \textbf{0.172} & 2.914 & 2.28 \\

Process-aware Reward  \citep{wang2024aligning}
& 0.486 & 0.682 & 0.781 
& 0.181 & 2.873 & \textbf{2.45} \\

\midrule
\textbf{MoRL Reward (Ours)} 
& 0.489 & \textbf{0.703} & \textbf{0.814}
& 0.179 & \textbf{2.756} & 2.42 \\
\bottomrule
\end{tabular}
}
\caption{Comparison of different reward designs on the CMS of HumanML3D. All methods share the same backbone and training setup, differing only in the reward used during motion generation.}
\label{tab:cms_reward}
\vspace{-0.5cm}
\end{table}

Outcome-based rewards (MotionR1-style) perform reasonably at R@1 but degrade at higher R-Precision, indicating omission of later-stage actions.
MotionRL improves realism (lower FID) but remains insensitive to stage-level semantic gaps.
The process-aware reward yields further gains by encouraging temporal coherence; yet, it still lacks fine-grained linguistic alignment.
In contrast, MoRL consistently improves R@2 and R@3 while achieving the lowest MM Distance, effectively reducing semantic drift in long-horizon sequences without sacrificing diversity.

\section{Conclusion}
We present MoRL, a unified multimodal motion model that integrates motion understanding and generation through reinforcement learning. With task-specific rewards and a COM decoding strategy, MoRL improves both logical consistency and perceptual realism. We also construct two large-scale synthetic CoT datasets for motion–language alignment. Experiments on HumanML3D and KIT-ML show that MoRL outperforms SOTA methods.

\section*{Limitations}
Our approach relies on rule-based reward design, which may require adaptation for new motion domains or styles. The Chain-of-Motion reasoning process introduces additional inference-time computation, limiting real-time applicability. Moreover, our method operates on discretized motion representations and does not explicitly model fine-grained contact dynamics or complex human–object interactions.

\bibliography{custom}

@article{wei2022chain,
  title={Chain-of-thought prompting elicits reasoning in large language models},
  author={Wei, Jason and Wang, Xuezhi and Schuurmans, Dale and Bosma, Maarten and Xia, Fei and Chi, Ed and Le, Quoc V and Zhou, Denny and others},
  journal={Advances in neural information processing systems},
  volume={35},
  pages={24824--24837},
  year={2022}
}

@article{zhang2023multimodal,
  title={Multimodal chain-of-thought reasoning in language models},
  author={Zhang, Zhuosheng and Zhang, Aston and Li, Mu and Zhao, Hai and Karypis, George and Smola, Alex},
  journal={arXiv preprint arXiv:2302.00923},
  year={2023}
}

@article{zhang2024improve,
  title={Improve vision language model chain-of-thought reasoning},
  author={Zhang, Ruohong and Zhang, Bowen and Li, Yanghao and Zhang, Haotian and Sun, Zhiqing and Gan, Zhe and Yang, Yinfei and Pang, Ruoming and Yang, Yiming},
  journal={arXiv preprint arXiv:2410.16198},
  year={2024}
}

@inproceedings{mitra2024compositional,
  title={Compositional chain-of-thought prompting for large multimodal models},
  author={Mitra, Chancharik and Huang, Brandon and Darrell, Trevor and Herzig, Roei},
  booktitle={Proceedings of the IEEE/CVF Conference on Computer Vision and Pattern Recognition},
  pages={14420--14431},
  year={2024}
}

@inproceedings{li2025lamp,
  title={LaMP: Language-Motion Pretraining for Motion Generation, Retrieval, and Captioning},
  author={Li, Zhe and Yuan, Weihao and Qiu, Lingteng and Zhu, Shenhao and Gu, Xiaodong and Shen, Weichao and Dong, Yuan and Dong, Zilong and Yang, Laurence Tianruo and others},
  booktitle={The Thirteenth International Conference on Learning Representations},
  year={2025}
}

@article{hao2024training,
  title={Training large language models to reason in a continuous latent space},
  author={Hao, Shibo and Sukhbaatar, Sainbayar and Su, DiJia and Li, Xian and Hu, Zhiting and Weston, Jason and Tian, Yuandong},
  journal={arXiv preprint arXiv:2412.06769},
  year={2024}
}

@article{yao2023tree,
  title={Tree of thoughts: Deliberate problem solving with large language models},
  author={Yao, Shunyu and Yu, Dian and Zhao, Jeffrey and Shafran, Izhak and Griffiths, Tom and Cao, Yuan and Narasimhan, Karthik},
  journal={Advances in neural information processing systems},
  volume={36},
  pages={11809--11822},
  year={2023}
}

@article{yuan2024advancing,
  title={Advancing llm reasoning generalists with preference trees},
  author={Yuan, Lifan and Cui, Ganqu and Wang, Hanbin and Ding, Ning and Wang, Xingyao and Deng, Jia and Shan, Boji and Chen, Huimin and Xie, Ruobing and Lin, Yankai and others},
  journal={arXiv preprint arXiv:2404.02078},
  year={2024}
}

@article{luan2024textcot,
  title={Textcot: Zoom in for enhanced multimodal text-rich image understanding},
  author={Luan, Bozhi and Feng, Hao and Chen, Hong and Wang, Yonghui and Zhou, Wengang and Li, Houqiang},
  journal={arXiv preprint arXiv:2404.09797},
  year={2024}
}

@article{yao2024mulberry,
  title={Mulberry: Empowering mllm with o1-like reasoning and reflection via collective monte carlo tree search},
  author={Yao, Huanjin and Huang, Jiaxing and Wu, Wenhao and Zhang, Jingyi and Wang, Yibo and Liu, Shunyu and Wang, Yingjie and Song, Yuxin and Feng, Haocheng and Shen, Li and others},
  journal={arXiv preprint arXiv:2412.18319},
  year={2024}
}

@article{thawakar2025llamav,
  title={Llamav-o1: Rethinking step-by-step visual reasoning in llms},
  author={Thawakar, Omkar and Dissanayake, Dinura and More, Ketan and Thawkar, Ritesh and Heakl, Ahmed and Ahsan, Noor and Li, Yuhao and Zumri, Mohammed and Lahoud, Jean and Anwer, Rao Muhammad and others},
  journal={arXiv preprint arXiv:2501.06186},
  year={2025}
}

@article{zhang2024llama,
  title={Llama-berry: Pairwise optimization for o1-like olympiad-level mathematical reasoning},
  author={Zhang, Di and Wu, Jianbo and Lei, Jingdi and Che, Tong and Li, Jiatong and Xie, Tong and Huang, Xiaoshui and Zhang, Shufei and Pavone, Marco and Li, Yuqiang and others},
  journal={arXiv preprint arXiv:2410.02884},
  year={2024}
}

@article{zhao2024marco,
  title={Marco-o1: Towards open reasoning models for open-ended solutions},
  author={Zhao, Yu and Yin, Huifeng and Zeng, Bo and Wang, Hao and Shi, Tianqi and Lyu, Chenyang and Wang, Longyue and Luo, Weihua and Zhang, Kaifu},
  journal={arXiv preprint arXiv:2411.14405},
  year={2024}
}

@article{guo2025deepseek,
  title={Deepseek-r1: Incentivizing reasoning capability in llms via reinforcement learning},
  author={Guo, Daya and Yang, Dejian and Zhang, Haowei and Song, Junxiao and Zhang, Ruoyu and Xu, Runxin and Zhu, Qihao and Ma, Shirong and Wang, Peiyi and Bi, Xiao and others},
  journal={arXiv preprint arXiv:2501.12948},
  year={2025}
}

@article{shao2024deepseekmath,
  title={Deepseekmath: Pushing the limits of mathematical reasoning in open language models},
  author={Shao, Zhihong and Wang, Peiyi and Zhu, Qihao and Xu, Runxin and Song, Junxiao and Bi, Xiao and Zhang, Haowei and Zhang, Mingchuan and Li, YK and Wu, Y and others},
  journal={arXiv preprint arXiv:2402.03300},
  year={2024}
}

@article{hu2025open,
  title={Open-Reasoner-Zero: An Open Source Approach to Scaling Up Reinforcement Learning on the Base Model},
  author={Hu, Jingcheng and Zhang, Yinmin and Han, Qi and Jiang, Daxin and Zhang, Xiangyu and Shum, Heung-Yeung},
  journal={arXiv preprint arXiv:2503.24290},
  year={2025}
}

@article{team2025kimi,
  title={Kimi k1. 5: Scaling reinforcement learning with llms},
  author={Team, Kimi and Du, Angang and Gao, Bofei and Xing, Bowei and Jiang, Changjiu and Chen, Cheng and Li, Cheng and Xiao, Chenjun and Du, Chenzhuang and Liao, Chonghua and others},
  journal={arXiv preprint arXiv:2501.12599},
  year={2025}
}

@inproceedings{goutsu2021linguistic,
  title={Linguistic descriptions of human motion with generative adversarial seq2seq learning},
  author={Goutsu, Yusuke and Inamura, Tetsunari},
  booktitle={2021 IEEE International conference on robotics and automation (ICRA)},
  pages={4281--4287},
  year={2021},
  organization={IEEE}
}

@article{yamada2018paired,
  title={Paired recurrent autoencoders for bidirectional translation between robot actions and linguistic descriptions},
  author={Yamada, Tatsuro and Matsunaga, Hiroyuki and Ogata, Tetsuya},
  journal={IEEE Robotics and Automation Letters},
  volume={3},
  number={4},
  pages={3441--3448},
  year={2018},
  publisher={IEEE}
}

@article{plappert2018learning,
  title={Learning a bidirectional mapping between human whole-body motion and natural language using deep recurrent neural networks},
  author={Plappert, Matthias and Mandery, Christian and Asfour, Tamim},
  journal={Robotics and Autonomous Systems},
  volume={109},
  pages={13--26},
  year={2018},
  publisher={Elsevier}
}

@inproceedings{guo2022tm2t,
  title={Tm2t: Stochastic and tokenized modeling for the reciprocal generation of 3d human motions and texts},
  author={Guo, Chuan and Zuo, Xinxin and Wang, Sen and Cheng, Li},
  booktitle={European Conference on Computer Vision},
  pages={580--597},
  year={2022},
  organization={Springer}
}

@article{radouane2024motion2language,
  title={Motion2language, unsupervised learning of synchronized semantic motion segmentation},
  author={Radouane, Karim and Tchechmedjiev, Andon and Lagarde, Julien and Ranwez, Sylvie},
  journal={Neural Computing and Applications},
  volume={36},
  number={8},
  pages={4401--4420},
  year={2024},
  publisher={Springer}
}

@inproceedings{zhang2024motion,
  title={Motion mamba: Efficient and long sequence motion generation},
  author={Zhang, Zeyu and Liu, Akide and Reid, Ian and Hartley, Richard and Zhuang, Bohan and Tang, Hao},
  booktitle={European Conference on Computer Vision},
  pages={265--282},
  year={2024},
  organization={Springer}
}

@article{zhang2024infinimotion,
  title={Infinimotion: Mamba boosts memory in transformer for arbitrary long motion generation},
  author={Zhang, Zeyu and Liu, Akide and Chen, Qi and Chen, Feng and Reid, Ian and Hartley, Richard and Zhuang, Bohan and Tang, Hao},
  journal={arXiv preprint arXiv:2407.10061},
  year={2024}
}

@article{zhang2025motion,
  title={Motion anything: Any to motion generation},
  author={Zhang, Zeyu and Wang, Yiran and Mao, Wei and Li, Danning and Zhao, Rui and Wu, Biao and Song, Zirui and Zhuang, Bohan and Reid, Ian and Hartley, Richard},
  journal={arXiv preprint arXiv:2503.06955},
  year={2025}
}

@article{zhang2024motionavatar,
  title={Motion avatar: Generate human and animal avatars with arbitrary motion},
  author={Zhang, Zeyu and Wang, Yiran and Wu, Biao and Chen, Shuo and Zhang, Zhiyuan and Huang, Shiya and Zhang, Wenbo and Fang, Meng and Chen, Ling and Zhao, Yang},
  journal={arXiv preprint arXiv:2405.11286},
  year={2024}
}

@article{zhang2024kmm,
  title={Kmm: Key frame mask mamba for extended motion generation},
  author={Zhang, Zeyu and Gao, Hang and Liu, Akide and Chen, Qi and Chen, Feng and Wang, Yiran and Li, Danning and Zhao, Rui and Li, Zhenming and Zhou, Zhongwen and others},
  journal={arXiv preprint arXiv:2411.06481},
  year={2024}
}

@inproceedings{zhang2025flashmo,
  title={Flashmo: Geometric interpolants and frequency-aware sparsity for scalable efficient motion generation},
  author={Zhang, Zeyu and Wang, Yiran and Li, Danning and Gong, Dong and Reid, Ian and Hartley, Richard},
  booktitle={The Thirty-ninth Annual Conference on Neural Information Processing Systems}
}

@article{wang2026safemo,
  title={SafeMo: Linguistically Grounded Unlearning for Trustworthy Text-to-Motion Generation},
  author={Wang, Yiling and Zhang, Zeyu and Wang, Yiran and Tang, Hao},
  journal={arXiv preprint arXiv:2601.00590},
  year={2026}
}

@article{radouane2023guided,
  title={Guided attention for interpretable motion captioning},
  author={Radouane, Karim and Lagarde, Julien and Ranwez, Sylvie and Tchechmedjiev, Andon},
  journal={arXiv preprint arXiv:2310.07324},
  year={2023}
}

@article{li2025remomask,
  title={Remomask: Retrieval-augmented masked motion generation},
  author={Li, Zhengdao and Wang, Siheng and Zhang, Zeyu and Tang, Hao},
  journal={arXiv preprint arXiv:2508.02605},
  year={2025}
}

@article{wang2024aligning,
  title={Aligning human motion generation with human perceptions},
  author={Wang, Haoru and Zhu, Wentao and Miao, Luyi and Xu, Yishu and Gao, Feng and Tian, Qi and Wang, Yizhou},
  journal={arXiv preprint arXiv:2407.02272},
  year={2024}
}

@inproceedings{zhou2024avatargpt,
  title={Avatargpt: All-in-one framework for motion understanding planning generation and beyond},
  author={Zhou, Zixiang and Wan, Yu and Wang, Baoyuan},
  booktitle={Proceedings of the IEEE/CVF Conference on Computer Vision and Pattern Recognition},
  pages={1357--1366},
  year={2024}
}

@inproceedings{zhang2024motiongpt,
  title={Motiongpt: Finetuned llms are general-purpose motion generators},
  author={Zhang, Yaqi and Huang, Di and Liu, Bin and Tang, Shixiang and Lu, Yan and Chen, Lu and Bai, Lei and Chu, Qi and Yu, Nenghai and Ouyang, Wanli},
  booktitle={Proceedings of the AAAI Conference on Artificial Intelligence},
  volume={38},
  number={7},
  pages={7368--7376},
  year={2024}
}

@inproceedings{ribeiro2024motiongpt,
  title={Motiongpt: Human motion synthesis with improved diversity and realism via gpt-3 prompting},
  author={Ribeiro-Gomes, Jose and Cai, Tianhui and Milacski, Zolt{\'a}n A and Wu, Chen and Prakash, Aayush and Takagi, Shingo and Aubel, Amaury and Kim, Daeil and Bernardino, Alexandre and De La Torre, Fernando},
  booktitle={Proceedings of the IEEE/CVF Winter Conference on Applications of Computer Vision},
  pages={5070--5080},
  year={2024}
}

@article{jiang2023motiongpt,
  title={Motiongpt: Human motion as a foreign language},
  author={Jiang, Biao and Chen, Xin and Liu, Wen and Yu, Jingyi and Yu, Gang and Chen, Tao},
  journal={Advances in Neural Information Processing Systems},
  volume={36},
  pages={20067--20079},
  year={2023}
}

@article{wang2024motiongpt,
  title={MotionGPT-2: A General-Purpose Motion-Language Model for Motion Generation and Understanding},
  author={Wang, Yuan and Huang, Di and Zhang, Yaqi and Ouyang, Wanli and Jiao, Jile and Feng, Xuetao and Zhou, Yan and Wan, Pengfei and Tang, Shixiang and Xu, Dan},
  journal={arXiv preprint arXiv:2410.21747},
  year={2024}
}

@article{wu2024motion,
  title={Motion-Agent: A Conversational Framework for Human Motion Generation with LLMs},
  author={Wu, Qi and Zhao, Yubo and Wang, Yifan and Liu, Xinhang and Tai, Yu-Wing and Tang, Chi-Keung},
  journal={arXiv preprint arXiv:2405.17013},
  year={2024}
}

@inproceedings{guo2022generating,
  title={Generating diverse and natural 3d human motions from text},
  author={Guo, Chuan and Zou, Shihao and Zuo, Xinxin and Wang, Sen and Ji, Wei and Li, Xingyu and Cheng, Li},
  booktitle={Proceedings of the IEEE/CVF conference on computer vision and pattern recognition},
  pages={5152--5161},
  year={2022}
}

@article{plappert2016kit,
  title={The kit motion-language dataset},
  author={Plappert, Matthias and Mandery, Christian and Asfour, Tamim},
  journal={Big data},
  volume={4},
  number={4},
  pages={236--252},
  year={2016},
  publisher={Mary Ann Liebert, Inc. 140 Huguenot Street, 3rd Floor New Rochelle, NY 10801 USA}
}

@inproceedings{jiang2024motionchain,
  title={Motionchain: Conversational motion controllers via multimodal prompts},
  author={Jiang, Biao and Chen, Xin and Zhang, Chi and Yin, Fukun and Li, Zhuoyuan and Yu, Gang and Fan, Jiayuan},
  booktitle={European Conference on Computer Vision},
  pages={54--74},
  year={2024},
  organization={Springer}
}

@inproceedings{lyu2025towards,
  title={Towards unified human motion-language understanding via sparse interpretable characterization},
  author={Lyu, Guangtao and Xu, Chenghao and Yan, Jiexi and Yang, Muli and Deng, Cheng},
  booktitle={The Thirteenth International Conference on Learning Representations},
  year={2025}
}

@article{ling2024motionllama,
  title={MotionLLaMA: A Unified Framework for Motion Synthesis and Comprehension},
  author={Ling, Zeyu and Han, Bo and Li, Shiyang and Shen, Hongdeng and Cheng, Jikang and Zou, Changqing},
  journal={arXiv e-prints},
  pages={arXiv--2411},
  year={2024}
}

@article{comanici2025gemini,
  title={Gemini 2.5: Pushing the frontier with advanced reasoning, multimodality, long context, and next generation agentic capabilities},
  author={Comanici, Gheorghe and Bieber, Eric and Schaekermann, Mike and Pasupat, Ice and Sachdeva, Noveen and Dhillon, Inderjit and Blistein, Marcel and Ram, Ori and Zhang, Dan and Rosen, Evan and others},
  journal={arXiv preprint arXiv:2507.06261},
  year={2025}
}

@incollection{loper2023smpl,
  title={SMPL: A skinned multi-person linear model},
  author={Loper, Matthew and Mahmood, Naureen and Romero, Javier and Pons-Moll, Gerard and Black, Michael J},
  booktitle={Seminal Graphics Papers: Pushing the Boundaries, Volume 2},
  pages={851--866},
  year={2023}
}

@inproceedings{pavlakos2019expressive,
  title={Expressive body capture: 3d hands, face, and body from a single image},
  author={Pavlakos, Georgios and Choutas, Vasileios and Ghorbani, Nima and Bolkart, Timo and Osman, Ahmed AA and Tzionas, Dimitrios and Black, Michael J},
  booktitle={Proceedings of the IEEE/CVF conference on computer vision and pattern recognition},
  pages={10975--10985},
  year={2019}
}

@article{zhang2023finemogen,
  title={Finemogen: Fine-grained spatio-temporal motion generation and editing},
  author={Zhang, Mingyuan and Li, Huirong and Cai, Zhongang and Ren, Jiawei and Yang, Lei and Liu, Ziwei},
  journal={Advances in Neural Information Processing Systems},
  volume={36},
  pages={13981--13992},
  year={2023}
}

@article{liu2024motionrl,
  title={Motionrl: Align text-to-motion generation to human preferences with multi-reward reinforcement learning},
  author={Liu, Xiaoyang and Mao, Yunyao and Zhou, Wengang and Li, Houqiang},
  journal={arXiv preprint arXiv:2410.06513},
  year={2024}
}

@article{ouyang2025motion,
  title={Motion-R1: Chain-of-Thought Reasoning and Reinforcement Learning for Human Motion Generation},
  author={Ouyang, Runqi and Li, Haoyun and Zhang, Zhenyuan and Wang, Xiaofeng and Zhu, Zheng and Huang, Guan and Wang, Xingang},
  journal={arXiv preprint arXiv:2506.10353},
  year={2025}
}

@article{chen2024motionllm,
  title={Motionllm: Understanding human behaviors from human motions and videos},
  author={Chen, Ling-Hao and Lu, Shunlin and Zeng, Ailing and Zhang, Hao and Wang, Benyou and Zhang, Ruimao and Zhang, Lei},
  journal={arXiv preprint arXiv:2405.20340},
  year={2024}
}

@inproceedings{feng2024chatpose,
  title={Chatpose: Chatting about 3d human pose},
  author={Feng, Yao and Lin, Jing and Dwivedi, Sai Kumar and Sun, Yu and Patel, Priyanka and Black, Michael J},
  booktitle={Proceedings of the IEEE/CVF conference on computer vision and pattern recognition},
  pages={2093--2103},
  year={2024}
}

@article{lin2024chathuman,
  title={Chathuman: Language-driven 3d human understanding with retrieval-augmented tool reasoning},
  author={Lin, Jing and Feng, Yao and Liu, Weiyang and Black, Michael J},
  journal={arXiv preprint arXiv:2405.04533},
  volume={2},
  year={2024}
}

@inproceedings{li2025unimotion,
  title={Unimotion: Unifying 3d human motion synthesis and understanding},
  author={Li, Chuqiao and Chibane, Julian and He, Yannan and Pearl, Naama and Geiger, Andreas and Pons-Moll, Gerard},
  booktitle={2025 International Conference on 3D Vision (3DV)},
  pages={240--249},
  year={2025},
  organization={IEEE}
}

@article{sun2024coma,
  title={Coma: Compositional human motion generation with multi-modal agents},
  author={Sun, Shanlin and De Araujo, Gabriel and Xu, Jiaqi and Zhou, Shenghan and Zhang, Hanwen and Huang, Ziheng and You, Chenyu and Xie, Xiaohui},
  journal={arXiv preprint arXiv:2412.07320},
  year={2024}
}

@inproceedings{yu2025remogpt,
  title={ReMoGPT: Part-Level Retrieval-Augmented Motion-Language Models},
  author={Yu, Qing and Tanaka, Mikihiro and Fujiwara, Kent},
  booktitle={Proceedings of the AAAI Conference on Artificial Intelligence},
  volume={39},
  number={9},
  pages={9635--9643},
  year={2025}
}

@article{wang2023t2m,
  title={T2m-hifigpt: generating high quality human motion from textual descriptions with residual discrete representations},
  author={Wang, Congyi},
  journal={arXiv preprint arXiv:2312.10628},
  year={2023}
}

@inproceedings{zhang2024motiondiffuse,
  title     = {MotionDiffuse: Text-driven human motion generation with diffusion model},
  author    = {Zhang, Mingyuan and Cai, Zhipeng and Pan, Liang et al.},
  booktitle = {TPAMI},
  year      = {2024}
}

@inproceedings{zhang2023t2mgpt,
  title     = {T2M-GPT: Generating human motion from textual descriptions with discrete representations},
  author    = {Zhang, Jie and Zhang, Yu et al.},
  booktitle = {CVPR},
  year      = {2023}
}

@inproceedings{zhang2023remodiffuse,
  title     = {ReMoDiffuse: Retrieval-Augmented Motion Diffusion Model},
  author    = {Zhang, Mingyuan and Guo, Xiaoyu et al.},
  booktitle = {ICCV},
  year      = {2023}
}

@inproceedings{ahuja2019language2pose,
  title={Language2Pose: Natural Language Grounded Pose Forecasting},
  author={Ahuja, Chaitanya and Morency, Louis-Philippe},
  booktitle={Proceedings of the IEEE/CVF International Conference on 3D Vision (3DV)},
  pages={719--728},
  year={2019}
}

@inproceedings{bhattacharya2021text2gestures,
  title={Text2Gestures: A Transformer-Based Network for Generating Emotive Body Gestures},
  author={Bhattacharya, Uttaran and Rewkowski, Nathaniel and Banerjee, Abhishek and Guhan, Prashanth and Bera, Aniket and Manocha, Dinesh},
  booktitle={IEEE Virtual Reality and 3D User Interfaces (VR)},
  pages={1--10},
  year={2021}
}

@inproceedings{tevet2023mdm,
  title={Human Motion Diffusion Model},
  author={Tevet, Guy and others},
  booktitle={ICLR},
  year={2023}
}

@article{guo2024momask,
  title={MoMask: Hierarchical Residual Quantization for Text-to-Motion Generation},
  author={Guo, Chenyang and others},
  journal={arXiv preprint arXiv:2401.08564},
  year={2024}
}

@article{yuan2024mogents,
  title={MoGenTS: Efficient Text-to-Motion Synthesis via Transformer Sampling},
  author={Yuan, Y. and others},
  journal={arXiv preprint arXiv:2403.06789},
  year={2024}
}

@article{yu2024remogpt,
  title={ReMoGPT: Retrieval-Augmented Motion-Language Model},
  author={Yu, T. and Tanaka, K. and Fujiwara, T.},
  journal={arXiv preprint arXiv:2402.04567},
  year={2024}
}

@article{liao2024rmd,
  title={RMD: Residual Motion Diffusion for Text-to-Motion Generation},
  author={Liao, J. and others},
  journal={arXiv preprint arXiv:2404.09876},
  year={2024}
}

@inproceedings{kalakonda2024morag,
  title={MoRAG-Diffuse: Motion Retrieval-Augmented Diffusion},
  author={Kalakonda, Sai Shashank and Maheshwari, H. and Sarvadevabhatla, R. K.},
  booktitle={ACM Multimedia},
  year={2024}
}

@article{yang2025qwen3,
  title={Qwen3 technical report},
  author={Yang, An and Li, Anfeng and Yang, Baosong and Zhang, Beichen and Hui, Binyuan and Zheng, Bo and Yu, Bowen and Gao, Chang and Huang, Chengen and Lv, Chenxu and others},
  journal={arXiv preprint arXiv:2505.09388},
  year={2025}
}

\clearpage
\appendix

\section{LLM Use Declaration}
Large Language Models (ChatGPT) were used exclusively to improve the clarity and fluency of English writing. They were not involved in research ideation, experimental design, data analysis, or interpretation. The authors take full responsibility for all content.

\section{More Implementation Details}
\subsection{NLI Model for the Reasoning-Coherence Reward}

This section clarifies the natural language inference (NLI) model used to compute the reasoning-coherence reward. Reasoning traces generated by the model are purely textual. Therefore, we evaluate their logical consistency with the predicted answer based on an NLI task. In particular, we adopt DeBERTa-v3-large-MNLI as $f_{\text{NLI}}$.

The model is kept frozen during reinforcement learning. This avoids reward drift and provides a stable, stationary reward signal. The input to $f_{\text{NLI}}$ is a pair of text sequences (reasoning trace and answer). No motion features are used. Motion--text alignment is handled by a separate reward. We take the softmax probability of the \emph{entailment} class as the coherence score and apply group-wise normalization.

Shown as Table \ref{tab:NLI}, larger NLI models produce more stable entailment probabilities. Smaller models show higher variance when processing long reasoning traces, which reduces RL stability. DeBERTa-v3-large achieves the most consistent reward signal and yields the strongest gains in both understanding and generation tasks.

We also tested a larger NLI model (DeBERTa-v3-XL). Despite its size, it did not improve reward stability or downstream performance. Larger model tended to be overconfident and reacted too strongly to the noise in model-generated CoT traces, which have flexible and imperfect linguistic structure. This sensitivity increased reward variance on long reasoning chains and introduced instability during RL updates, ultimately reducing motion quality. The larger model also incurred significantly higher computational cost. In contrast, DeBERTa-v3-large provides the best trade-off between stability, robustness, and efficiency. We therefore use it as the default $f_{\text{NLI}}$ and keep it frozen throughout training to ensure reproducibility.

\begin{table}[t]
\centering
\small
\setlength{\tabcolsep}{4pt}
\begin{tabular}{lcccc}
\toprule
NLI Model & Params & Var.$\downarrow$ & BLEU$\uparrow$ & FID$\downarrow$ \\
\midrule
DeBERTa-v3-XL & 1.2B & Low–Med. & +3.3 & -2.0 \\
\textbf{DeBERTa-v3-L} & 435M & \textbf{Lowest} & \textbf{+3.4} & \textbf{-2.1} \\
RoBERTa-L & 355M & Low & +2.8 & -1.5 \\
DeBERTa-B & 110M & Med. & +1.9 & -0.8 \\
BERT-B & 110M & High & +0.7 & -0.2 \\
\bottomrule
\end{tabular}
\caption{Comparison of NLI models used as $f_{\text{NLI}}$.
Lower reward variance indicates more stable RL updates. BLEU↑ and FID↓ reflect performance gains relative to the SFT baseline.}
\label{tab:NLI}
\end{table}

\subsection{Reward Normalization}
\label{appendix:normalization}

Our reinforcement learning stage combines four heterogeneous reward components:
semantic alignment, reasoning coherence, physical plausibility, and text--motion consistency. 
These rewards naturally exhibit different dynamic ranges and variances, which may cause unstable optimization when used directly. 
To address this, we apply group-wise normalization to each reward component within a GRPO candidate group.

\paragraph{Group-wise Normalization.}
Given a candidate group $\{r_{1}, r_{2}, \ldots, r_{K}\}$, we normalize each reward as
\begin{equation}
\begin{aligned}
\tilde{r}_{i} &= \frac{r_{i} - \mu_{r}}{\sigma_{r} + \epsilon}, \\
\mu_{r} &= \frac{1}{K}\sum_{j=1}^{K} r_{j}, \\
\sigma_{r}^{2} &= \frac{1}{K}\sum_{j=1}^{K} (r_{j}-\mu_{r})^{2}.
\end{aligned}
\end{equation}
This ensures that rewards are centered and variance-controlled inside each GRPO update, 
which stabilizes the advantage computation and reduces gradient variance.

\paragraph{Component-wise Scaling.}
After normalization, we apply scalar weights $\lambda_{1}$ and $\lambda_{2}$ to the two physical plausibility rewards:
\begin{equation}
\begin{split}
R_{\text{phys}} &= - \lambda_1 \cdot L_{\text{joint}}(\hat{m}) 
- \lambda_2 \cdot L_{\text{vel}}(\hat{m}), \\
\lambda_{1} &= 0.8, \quad
\lambda_{2} = 0.2.
\end{split}
\end{equation}
These values are selected to balance the magnitude of joint-limit penalties and velocity-smoothness penalties.

\section{Inference Latency and Throughput of CoM}
\label{appendix:com_inference}

We report the end-to-end inference cost of CoM decoding and compare it with standard single-pass decoding in Table \ref{ete}. All measurements are obtained on a single NVIDIA A100 GPU using batch size 32 under the HumanML3D generation setting.

\begin{table}[t]
\centering
\small
\setlength{\tabcolsep}{4pt}
\begin{tabular}{lccc}
\toprule
Method & Lat. (ms)$\downarrow$ & Thru. (sps)$\uparrow$ & Cost \\
\midrule
Single-pass & 8.7 & 115 & baseline \\
CoM (K=8, T=2) & 18.4 & 55 & 2.1$\times$ \\
\bottomrule
\end{tabular}
\caption{End-to-end inference efficiency of single-pass decoding vs.\ CoM.
Latency (Lat.) is measured per sample. Throughput (Thru.) is computed at batch size 32.}
\label{ete}
\end{table}

CoM introduces moderate overhead because it evaluates multiple candidate trajectories during inference. 
However, candidate sampling is executed in parallel, so the cost scales sub-linearly with $K \times T$. 
Despite the increased latency, CoM consistently improves semantic alignment, reasoning coherence, and physical plausibility, making the additional cost acceptable for practical use.

\section{Choice of RL Optimizer}
To examine the effect of different reinforcement learning optimizers, we trained MoRL using PPO, DPO, and GRPO under identical settings. The results are summarized in Table~\ref{tab:rl_opt_compare}. GRPO provides the most stable and effective optimization for motion reasoning and generation.

PPO shows instability when handling long-horizon and multi-component rewards, leading to higher variance and slower convergence. DPO underperforms because it is designed for preference-based objectives and cannot fully exploit our structured reward components (semantic, reasoning, physical, and text–motion alignment). In contrast, GRPO stabilizes credit assignment over long reasoning chains and supports continuous reward shaping, resulting in consistent improvements across both understanding and generation metrics.

Overall, GRPO achieves the best balance of performance, training stability, and efficiency, and we therefore adopt it as our default RL optimizer.

\begin{table}[t]
\centering
\small
\setlength{\tabcolsep}{3pt}
\begin{tabular}{lcccccc}
\toprule
Method & BERT$\uparrow$ & CIDEr$\uparrow$ & R@1$\uparrow$ & FID$\downarrow$ & Var$\downarrow$ & Time$\downarrow$ \\
\midrule
SFT only & 42.65 & 33.88 & 0.420 & 0.232 & -- & -- \\
PPO & 45.12 & 34.95 & 0.492 & 0.228 & 1.00 & 1.35$\times$ \\
DPO & 44.31 & 34.10 & 0.468 & 0.241 & 0.87 & 1.20$\times$ \\
\textbf{GRPO} & \textbf{46.80} & \textbf{35.80} & \textbf{0.527} & \textbf{0.203} & \textbf{0.63} & \textbf{1.00$\times$} \\
\bottomrule
\end{tabular}
\caption{Comparison of different optimization strategies under identical settings. GRPO provides the best overall performance and training stability.}
\label{tab:rl_opt_compare}
\end{table}

\section{User Study}
\begin{figure}[t]
  \centering
\includegraphics[width=0.48\textwidth]{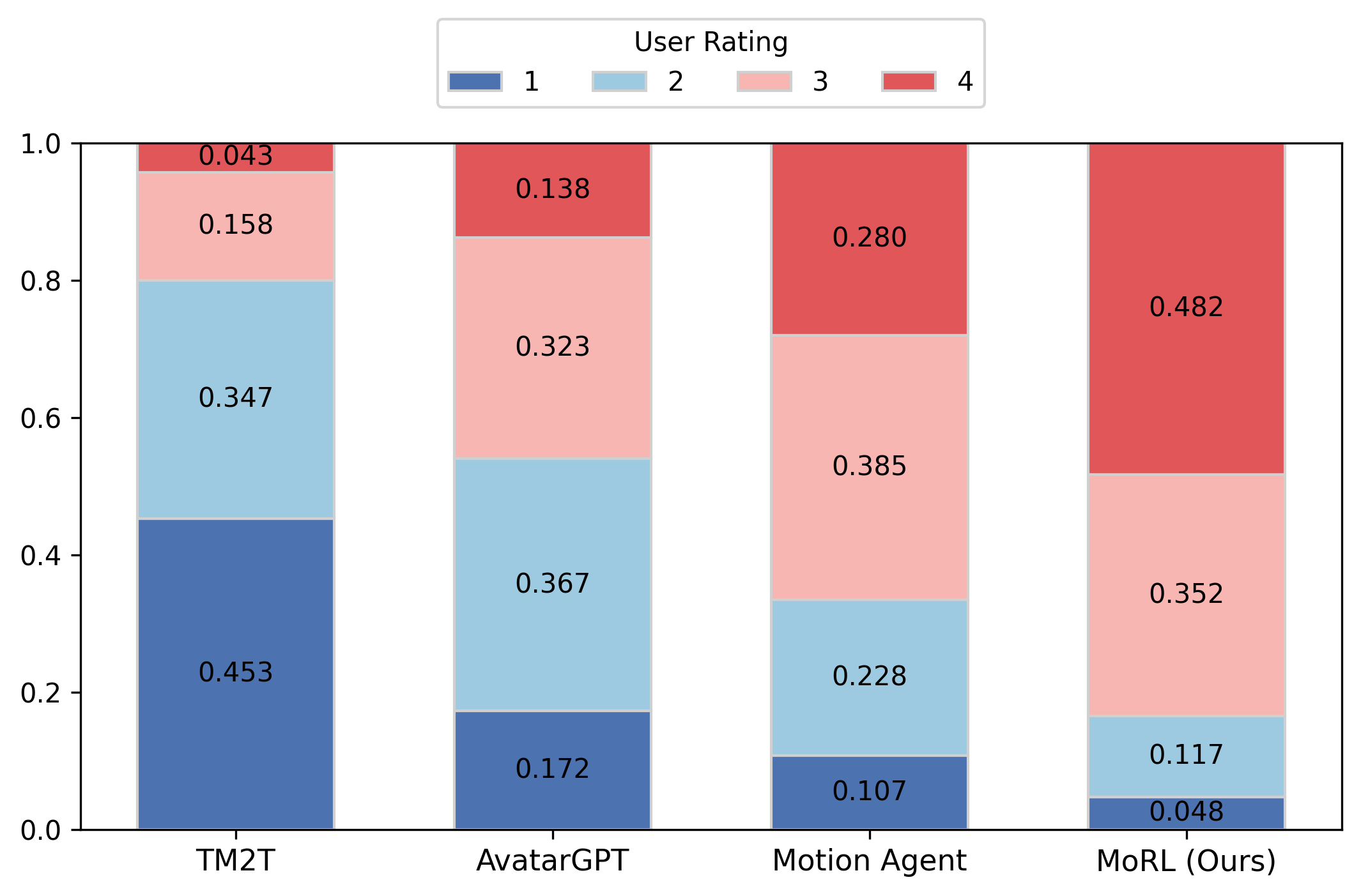} 
  \caption{Results of user study.}
  \label{fig:userstudy}
\end{figure}

We conduct a user study to evaluate text-to-motion generation from a human-centered perspective. We select 20 text prompts from the HumanML3D dataset and compare the motions generated by four methods: TM2T, AvatarGPT, Motion Agent, and our approach. A total of 20 participants are recruited for the evaluation. For each sample, participants are asked to assess the generated motions of all four methods using a four-point rating scale and to rank the methods from best to worst. During the evaluation, participants are instructed to focus on three key aspects: physical plausibility, motion smoothness, and semantic consistency between the generated motion and the input text. Figure~\ref{fig:userstudy} shows the distribution of user ratings for all compared methods. 

\begin{table*}[t]
\centering
\setlength{\tabcolsep}{2.5pt}
\renewcommand{\arraystretch}{1.0}

\begin{tabular}{@{}%
C{3.9cm}
C{5.3cm}
C{5.3cm}
@{}}
\toprule
\textbf{Text Prompt} &
\textbf{Motion Agent} &
\textbf{MoRL (Ours)} \\
\midrule

A person looks to the left then kicks something with their right foot.
&
\includegraphics[width=5.0cm,height=3.2cm,keepaspectratio]{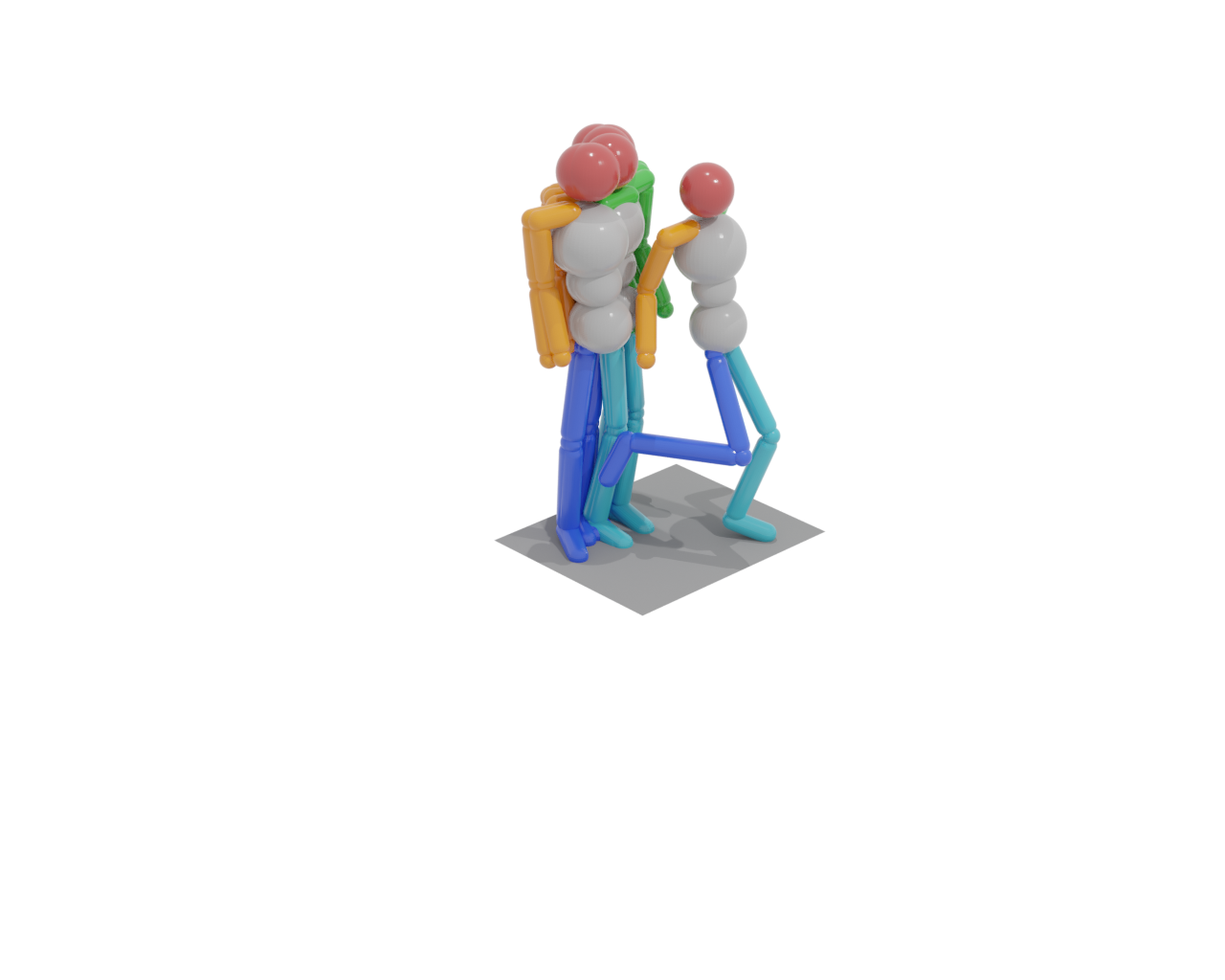}
&
\includegraphics[width=5.0cm,height=3.2cm,keepaspectratio]{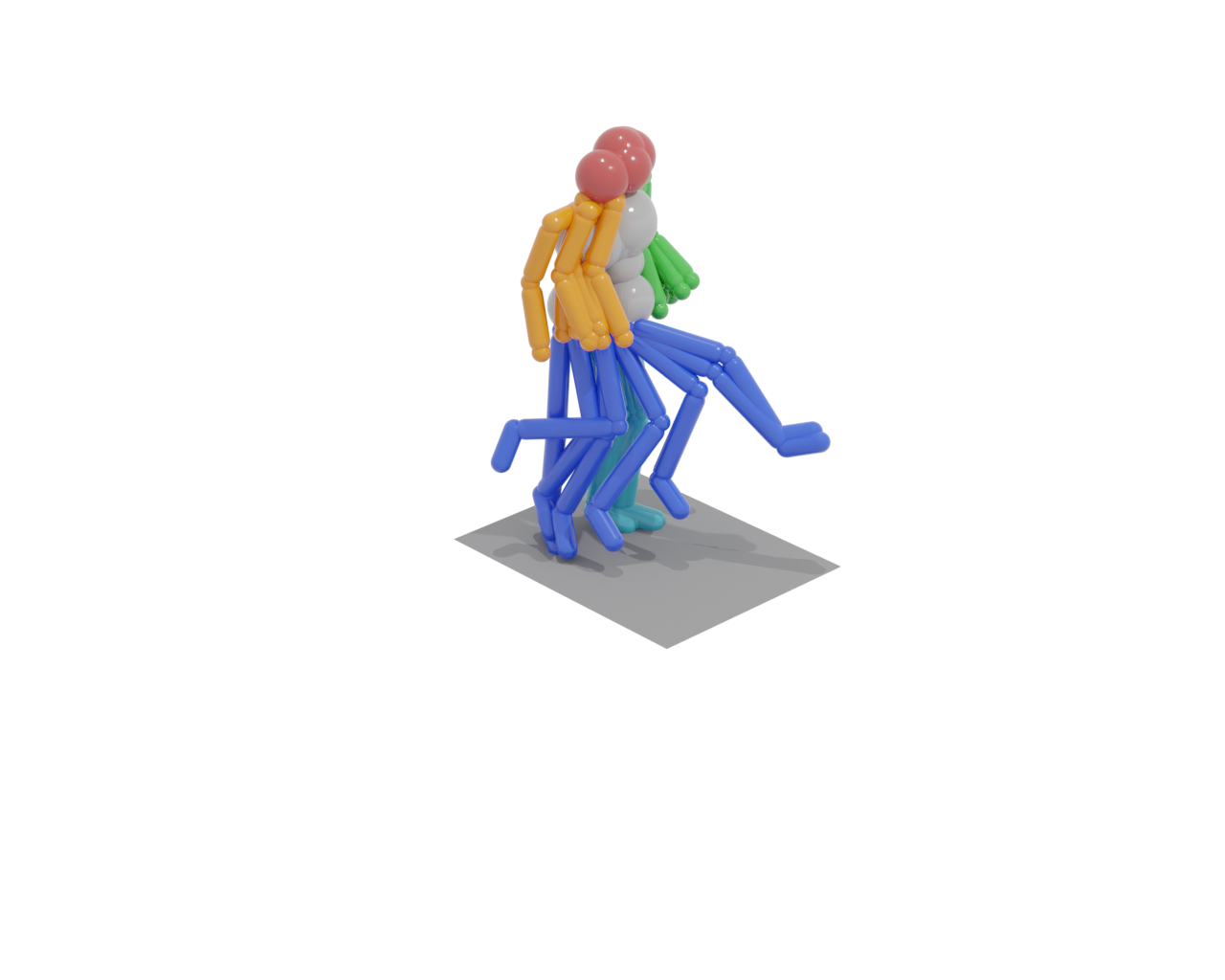}
\\

A person walks up stairs.
&
\includegraphics[width=5.0cm,height=3.2cm,keepaspectratio]{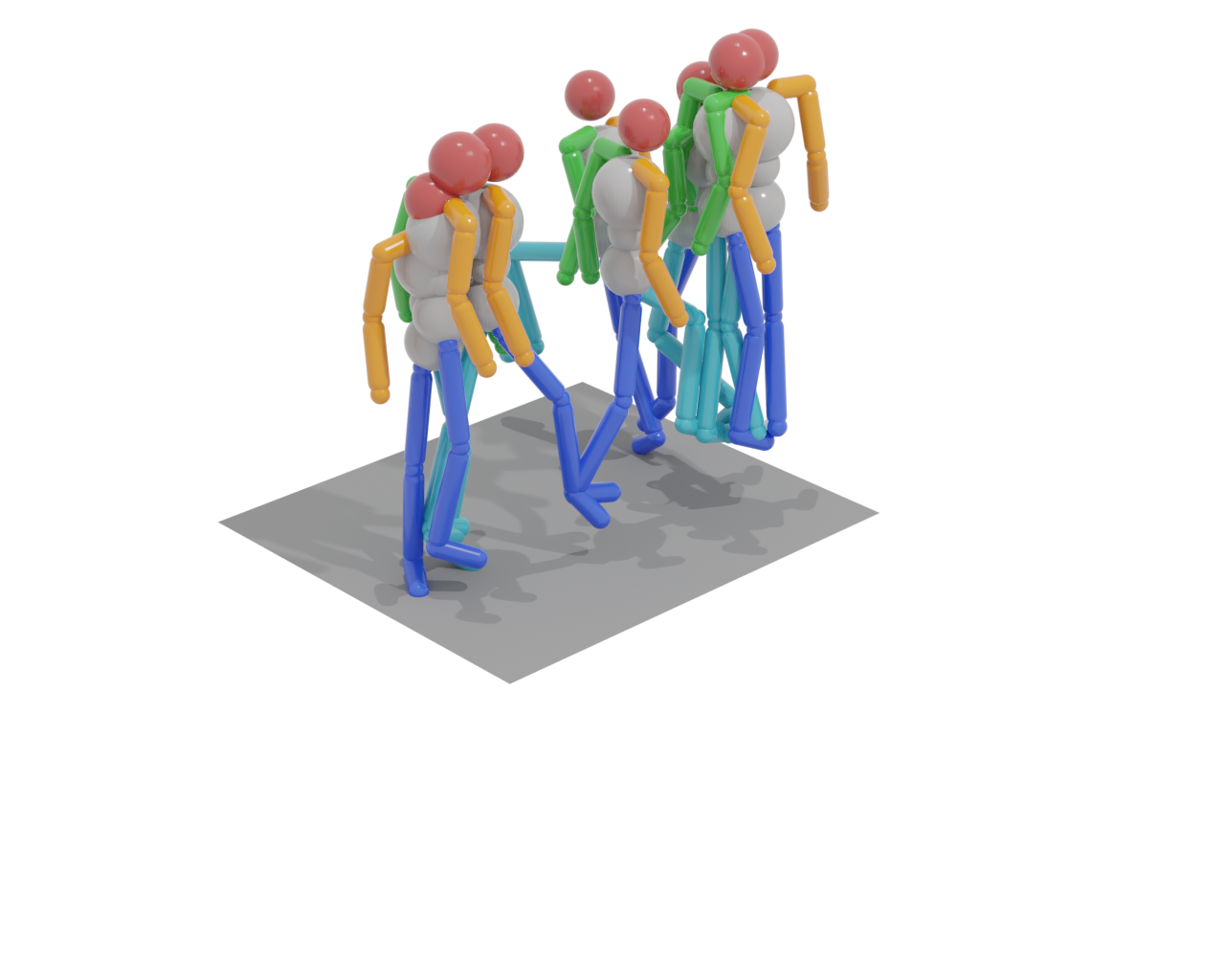}
&
\includegraphics[width=5.0cm,height=3.2cm,keepaspectratio]{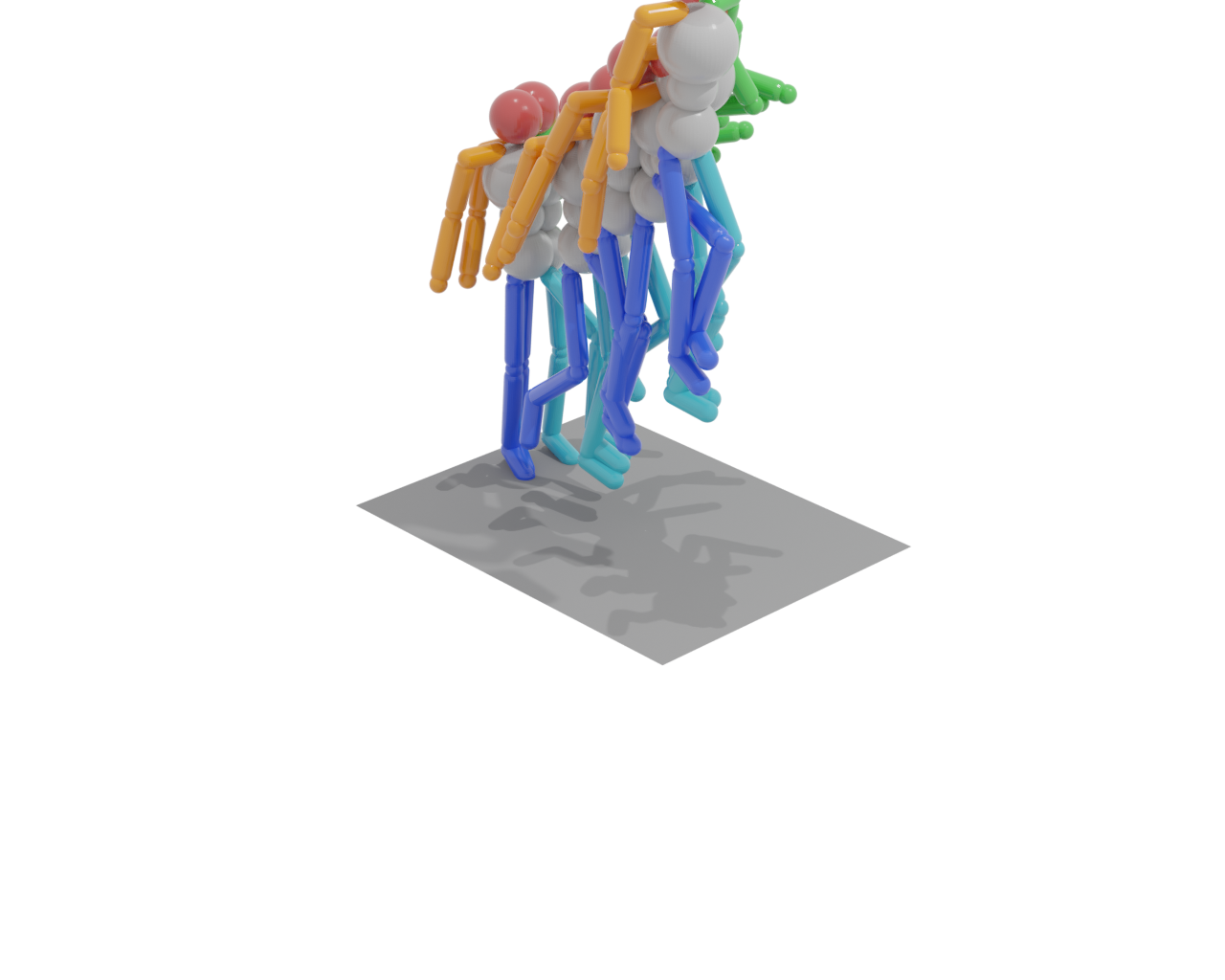}
\\

A person walks forward, slightly shifting to the right.
&
\includegraphics[width=5.0cm,height=3.2cm,keepaspectratio]{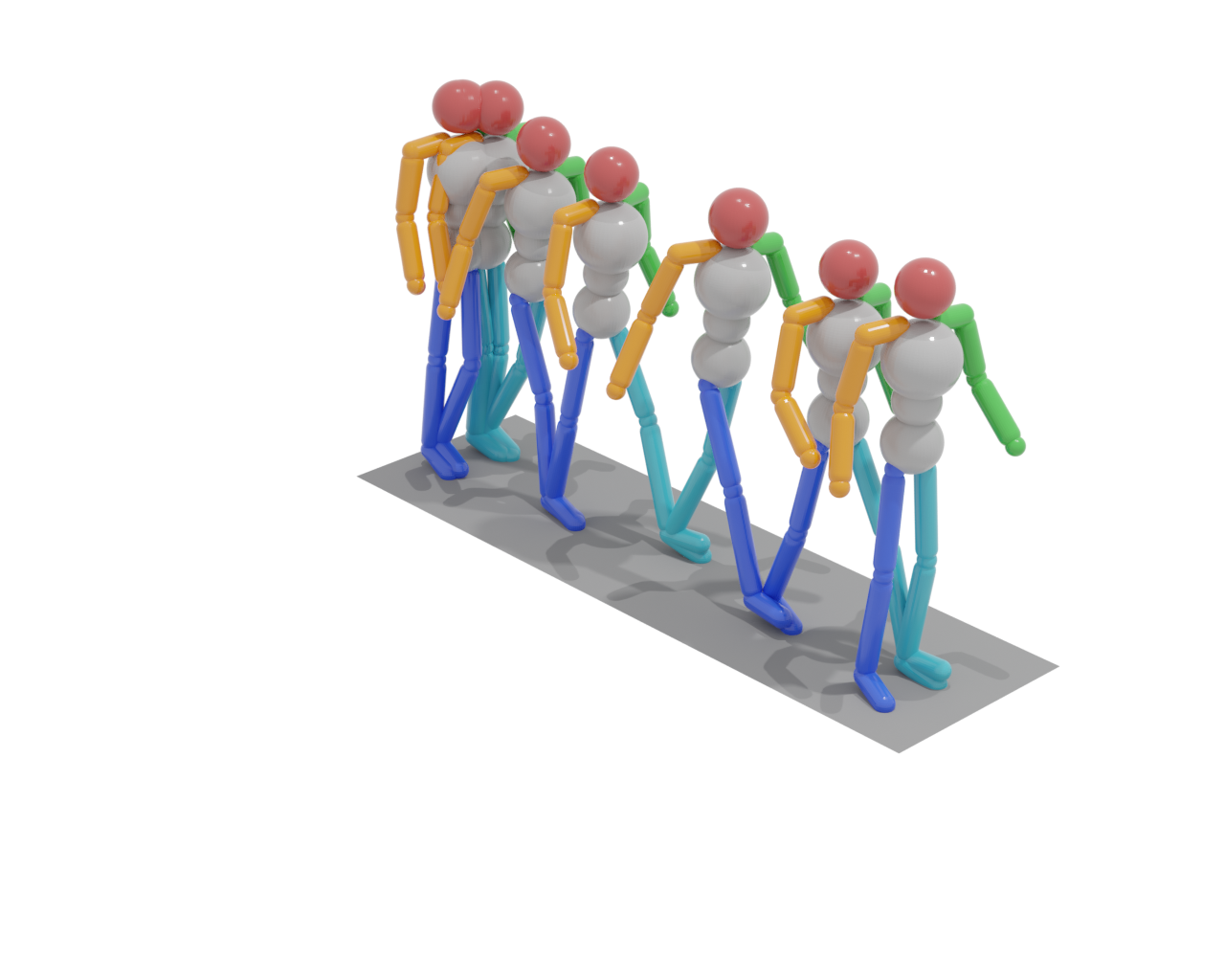}
&
\includegraphics[width=5.0cm,height=3.2cm,keepaspectratio]{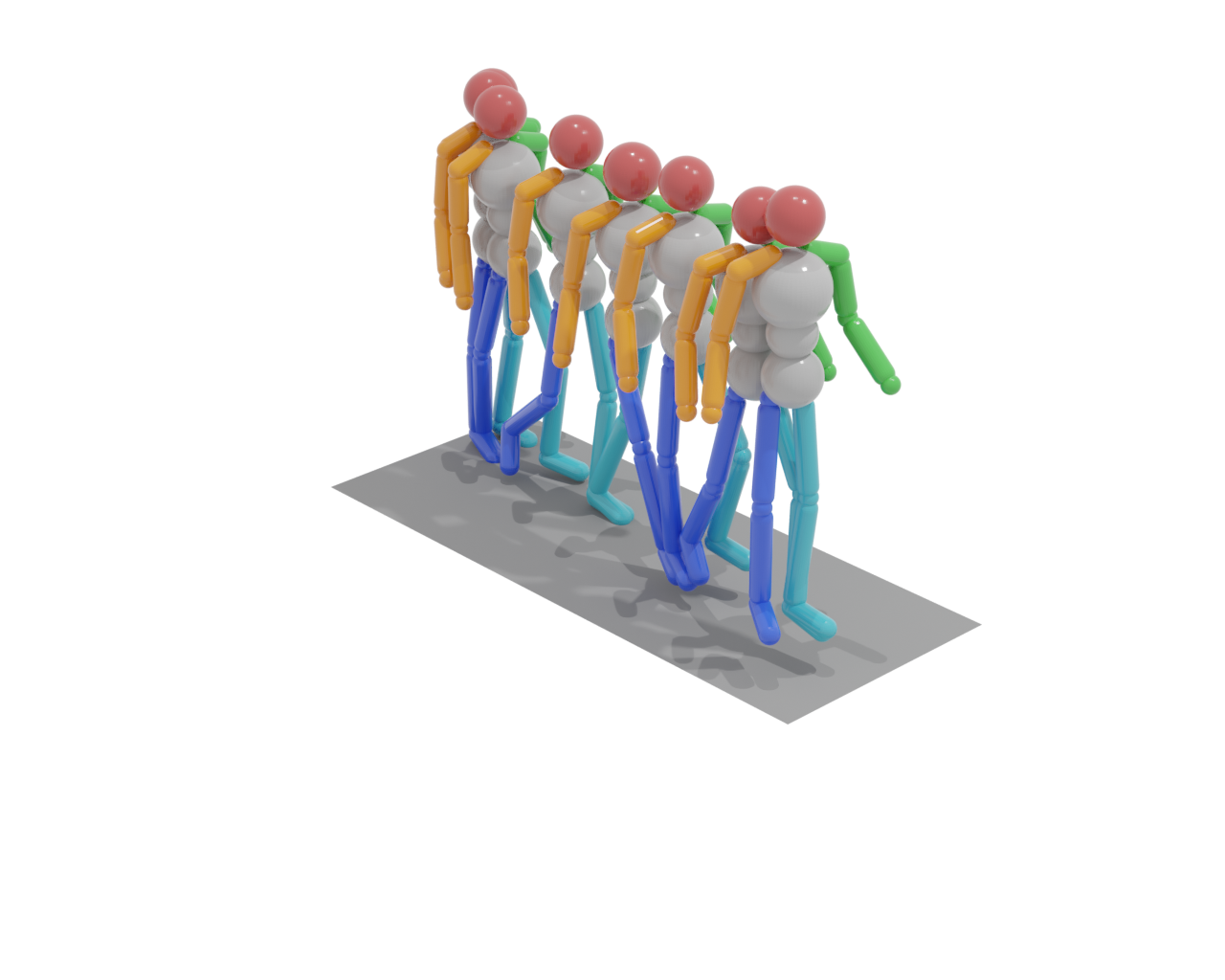}
\\

A person walks forward with a side-to-side sway.
&
\includegraphics[width=5.0cm,height=3.2cm,keepaspectratio]{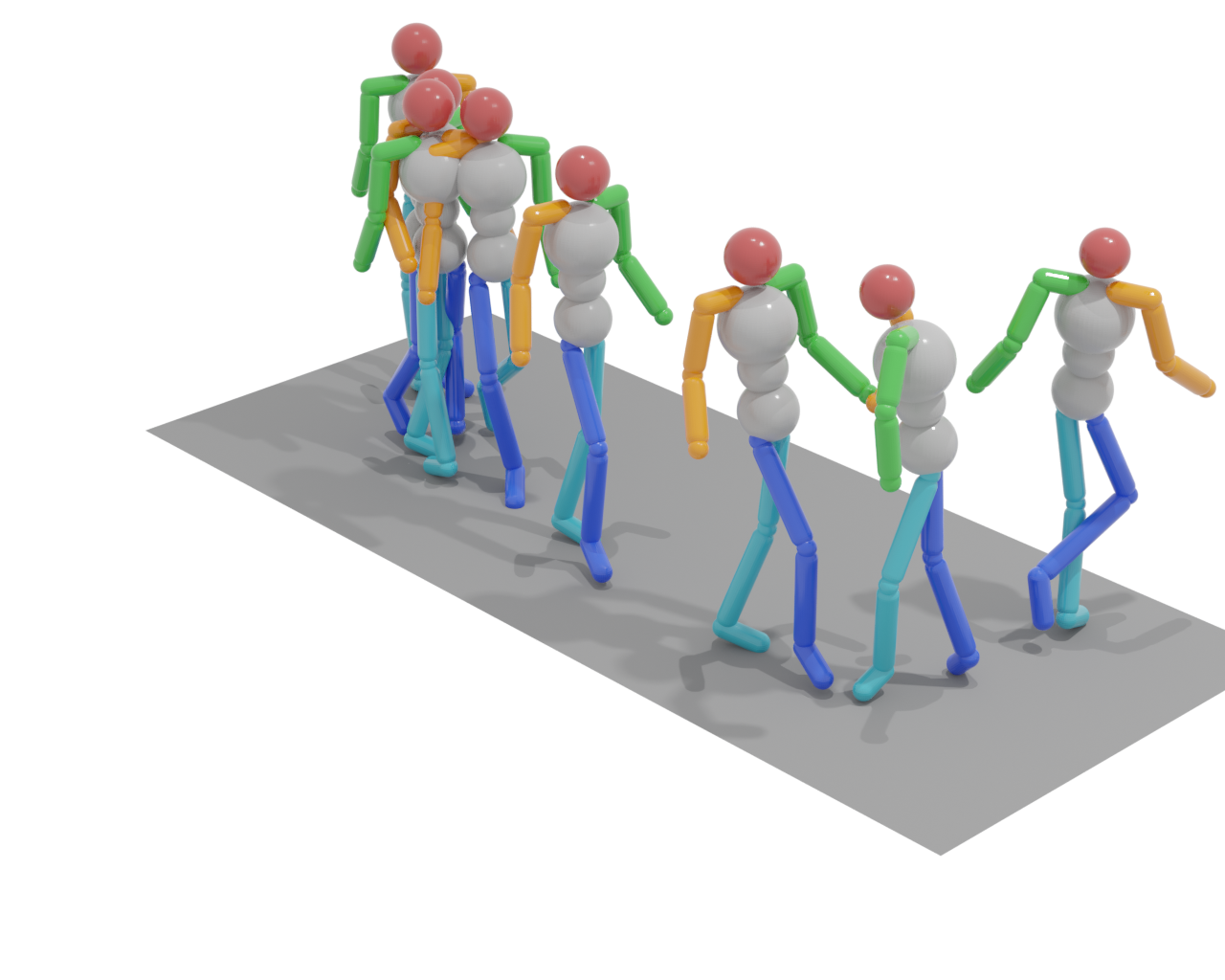}
&
\includegraphics[width=5.0cm,height=3.2cm,keepaspectratio]{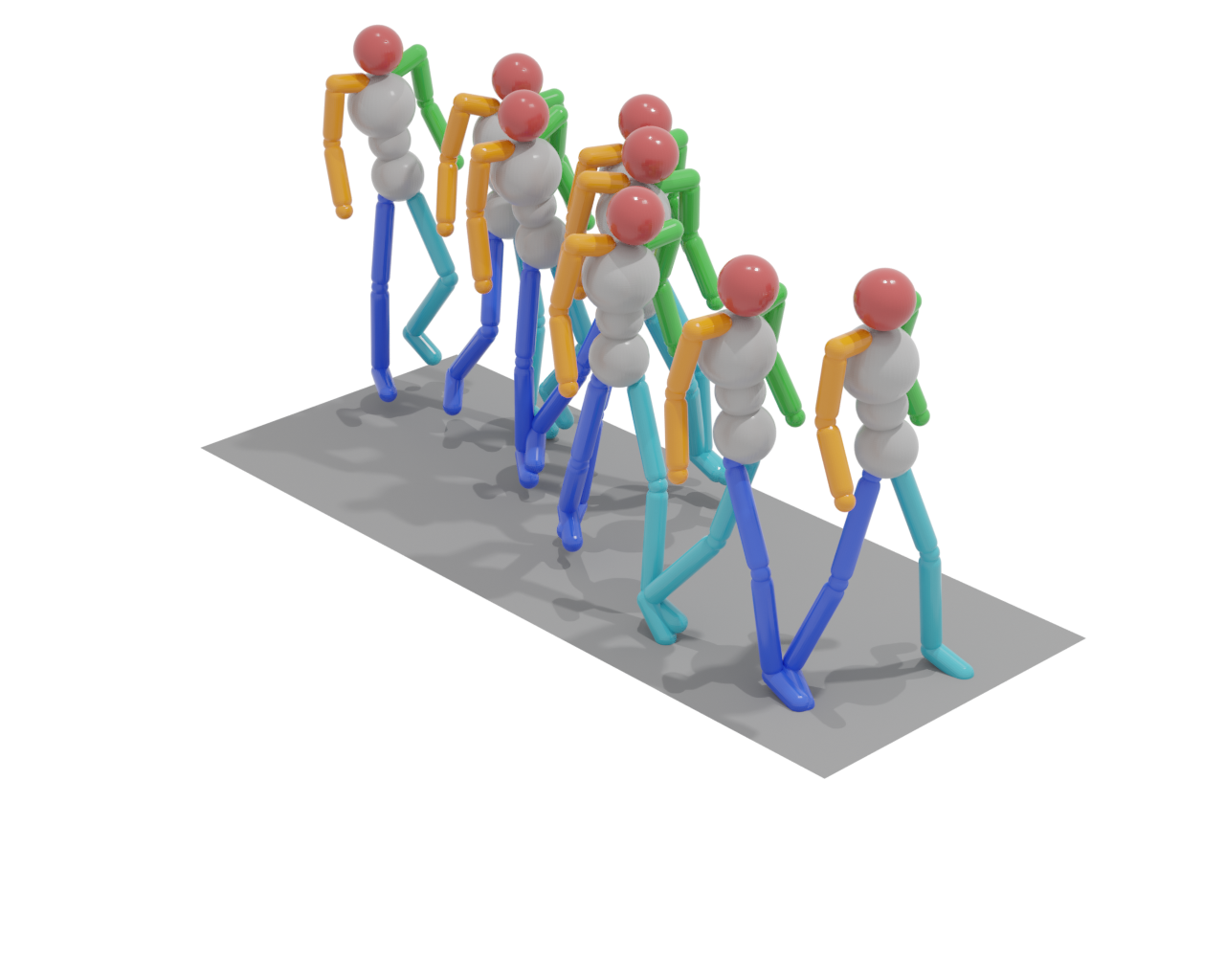}
\\

\bottomrule
\end{tabular}

\caption{Qualitative comparison (Part I).}
\label{tab:safemo_qualitative_simple_1}
\vspace{-0.8em}
\end{table*}

TM2T tends to receive lower ratings overall, which can be attributed to its limited capability in modeling long-term motion dynamics and complex text–motion relations. Despite this, TM2T still produces reasonable motions in simpler cases, reflecting the effectiveness of its early text-to-motion formulation. AvatarGPT and Motion Agent exhibit more balanced rating distributions, with a noticeable shift toward higher scores. This suggests improved motion quality and semantic alignment compared to earlier methods, although occasional low ratings indicate challenges in maintaining global motion coherence and physical stability over longer sequences. Our method demonstrates a clear concentration of high ratings and very few low-rated cases. This improvement is mainly due to the integration of structured motion composition, which facilitates coherent temporal transitions, and an explicit physical plausibility reward that helps suppress unrealistic poses and abrupt motion artifacts. As a result, participants consistently prefer our generated motions in terms of physical plausibility, motion smoothness, and semantic consistency.

\begin{table*}[t]
\centering
\setlength{\tabcolsep}{2.5pt}
\renewcommand{\arraystretch}{1.0}

\begin{tabular}{@{}%
C{3.9cm}
C{5.3cm}
C{5.3cm}
@{}}
\toprule
\textbf{Text Prompt} &
\textbf{Motion Agent} &
\textbf{MoRL (Ours)} \\
\midrule

Walking slowly along the path shaped like an infinity symbol.
&
\includegraphics[width=5.0cm,height=3.2cm,keepaspectratio]{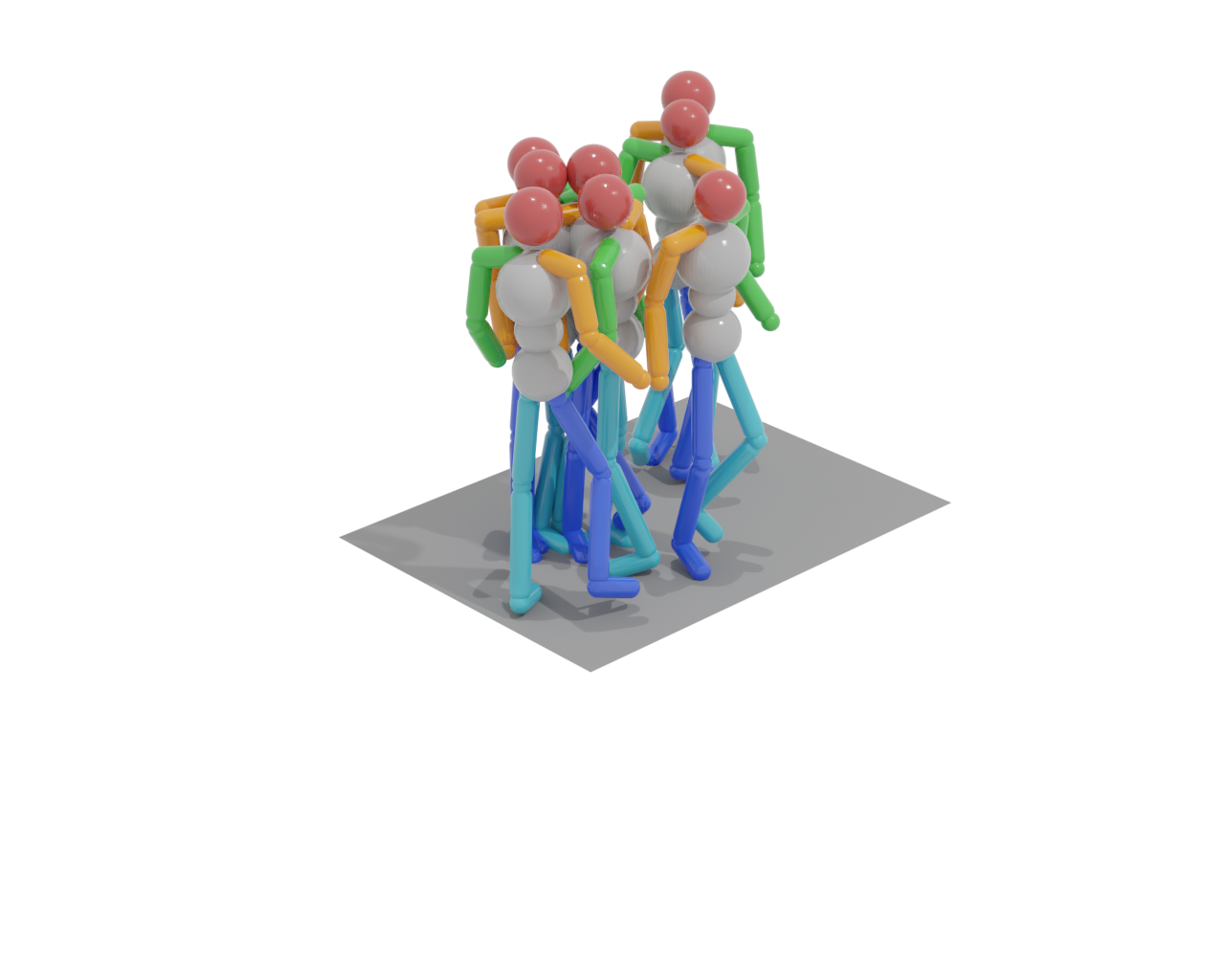}
&
\includegraphics[width=5.0cm,height=3.2cm,keepaspectratio]{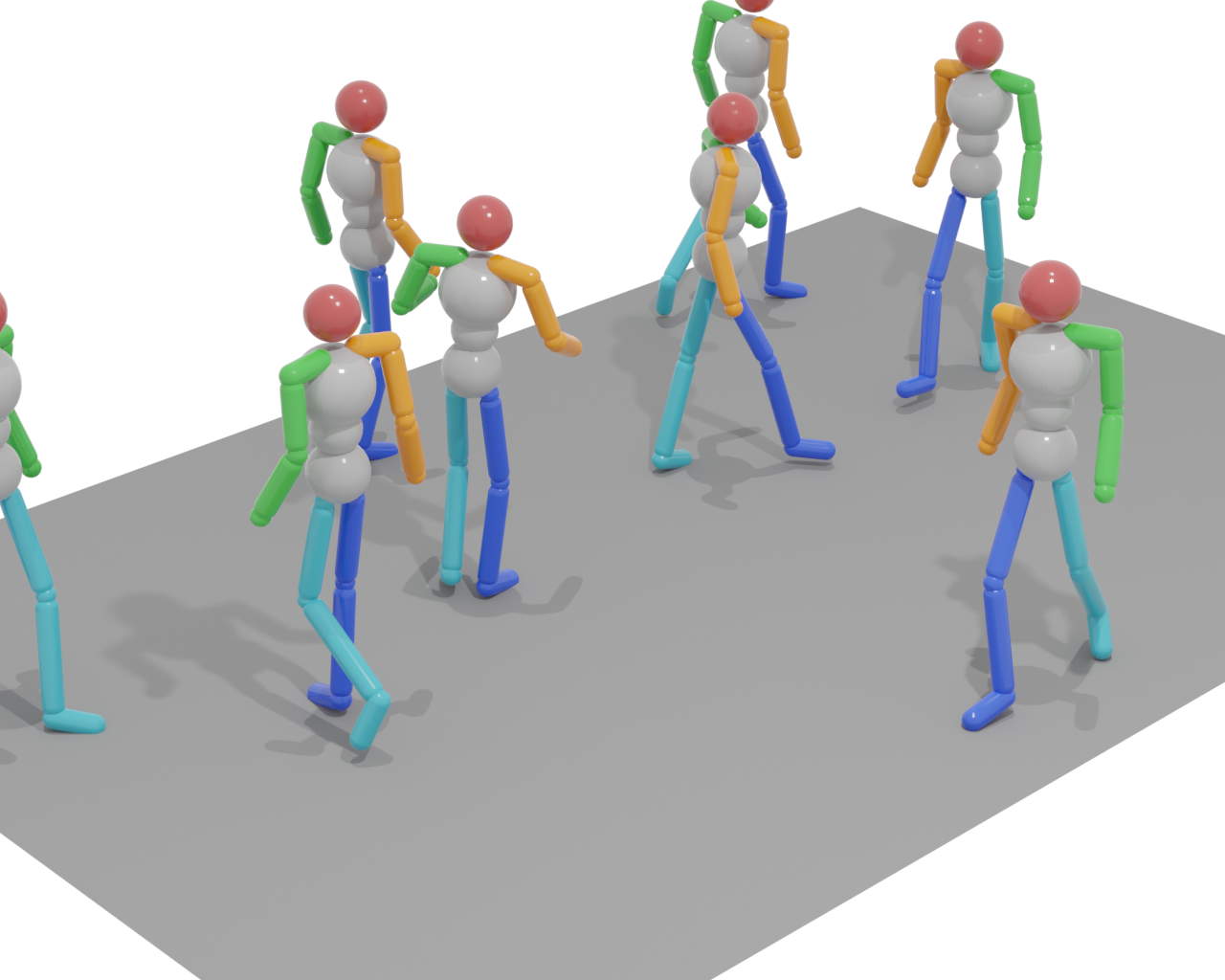}
\\

A person walks along a curved path to the right.
&
\includegraphics[width=5.0cm,height=3.2cm,keepaspectratio]{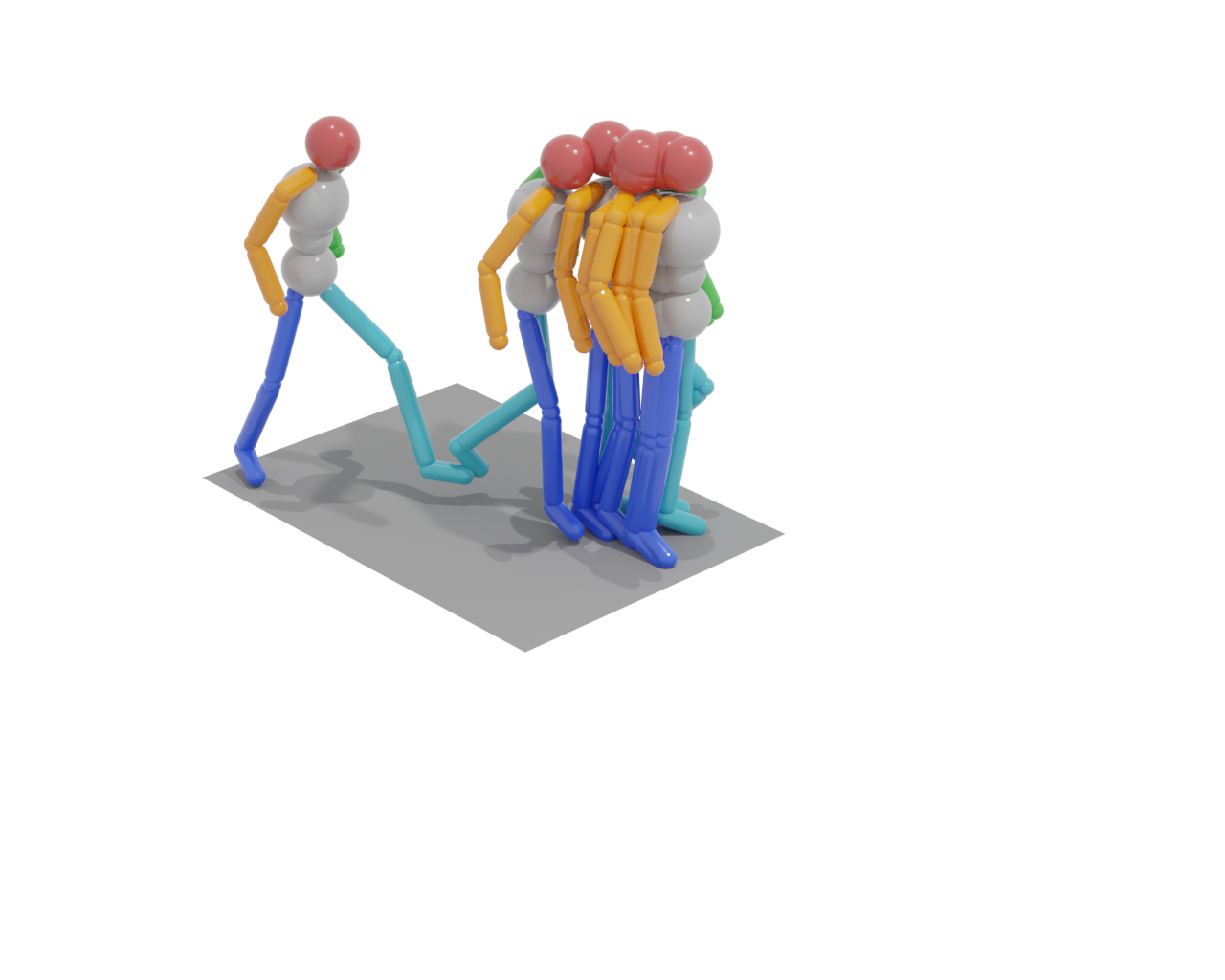}
&
\includegraphics[width=5.0cm,height=3.2cm,keepaspectratio]{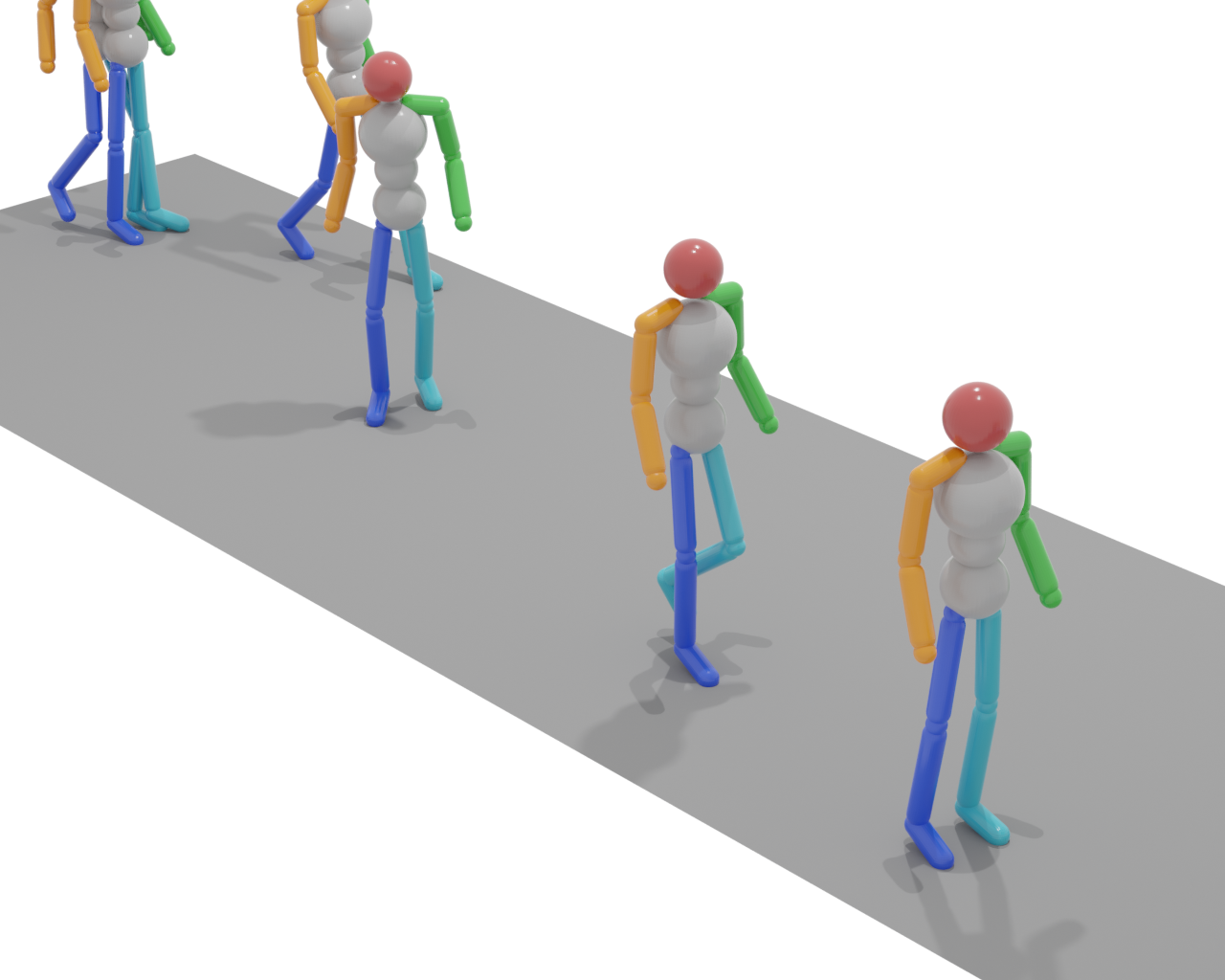}
\\

A person backflips three times in a row.
&
\includegraphics[width=5.0cm,height=3.2cm,keepaspectratio]{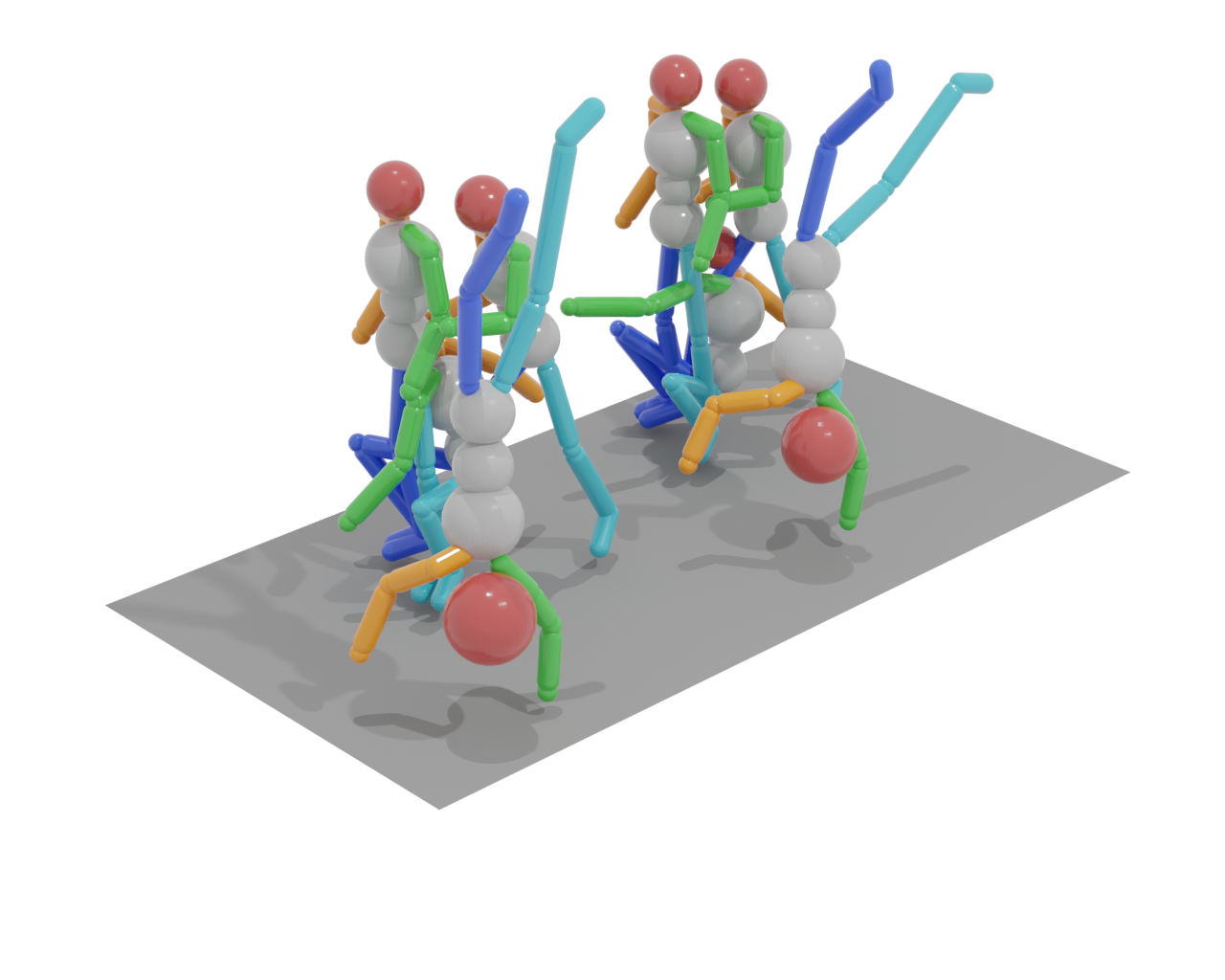}
&
\includegraphics[width=5.0cm,height=3.2cm,keepaspectratio]{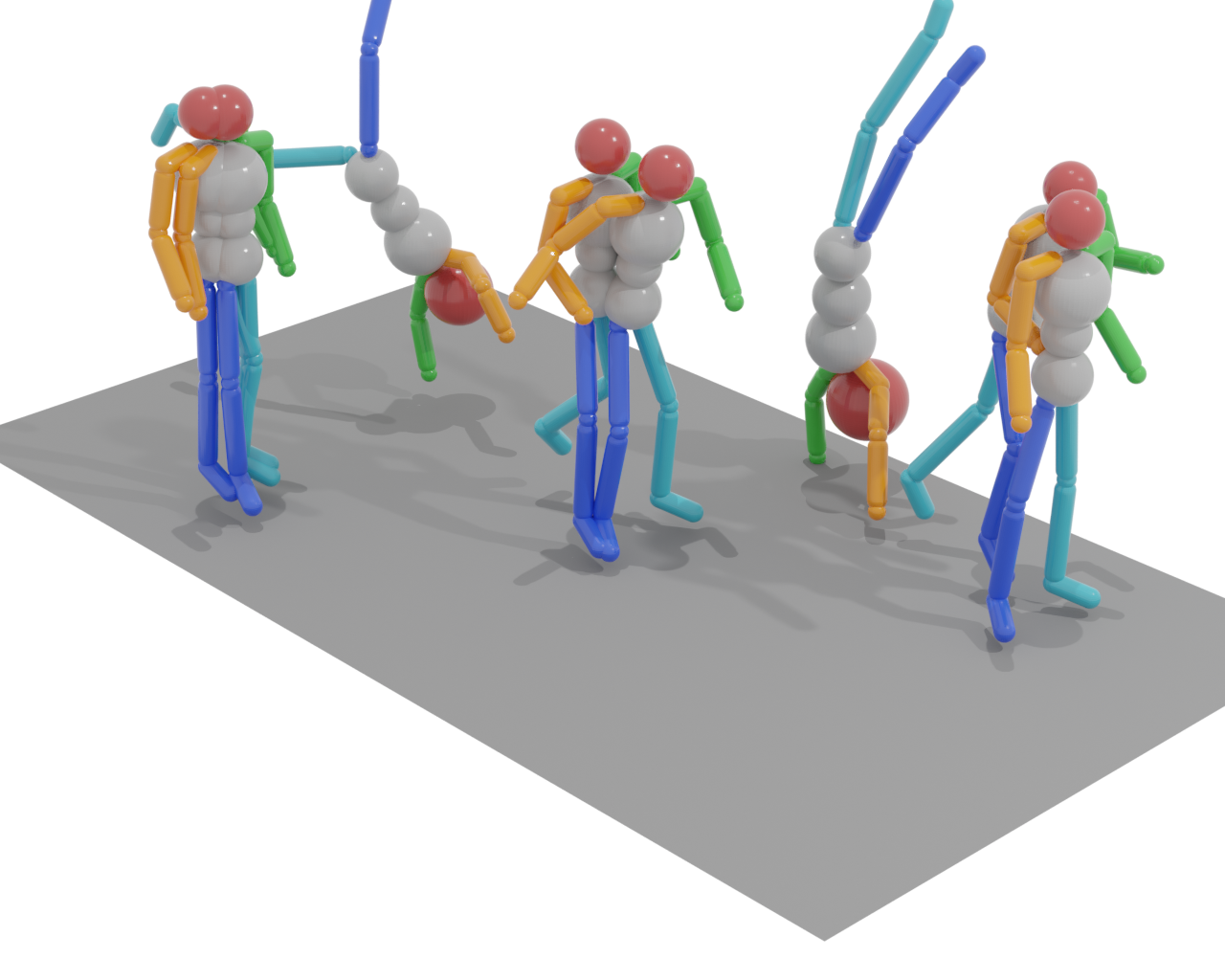}
\\

A person is practicing karate moves across the floor.
&
\includegraphics[width=5.0cm,height=3.2cm,keepaspectratio]{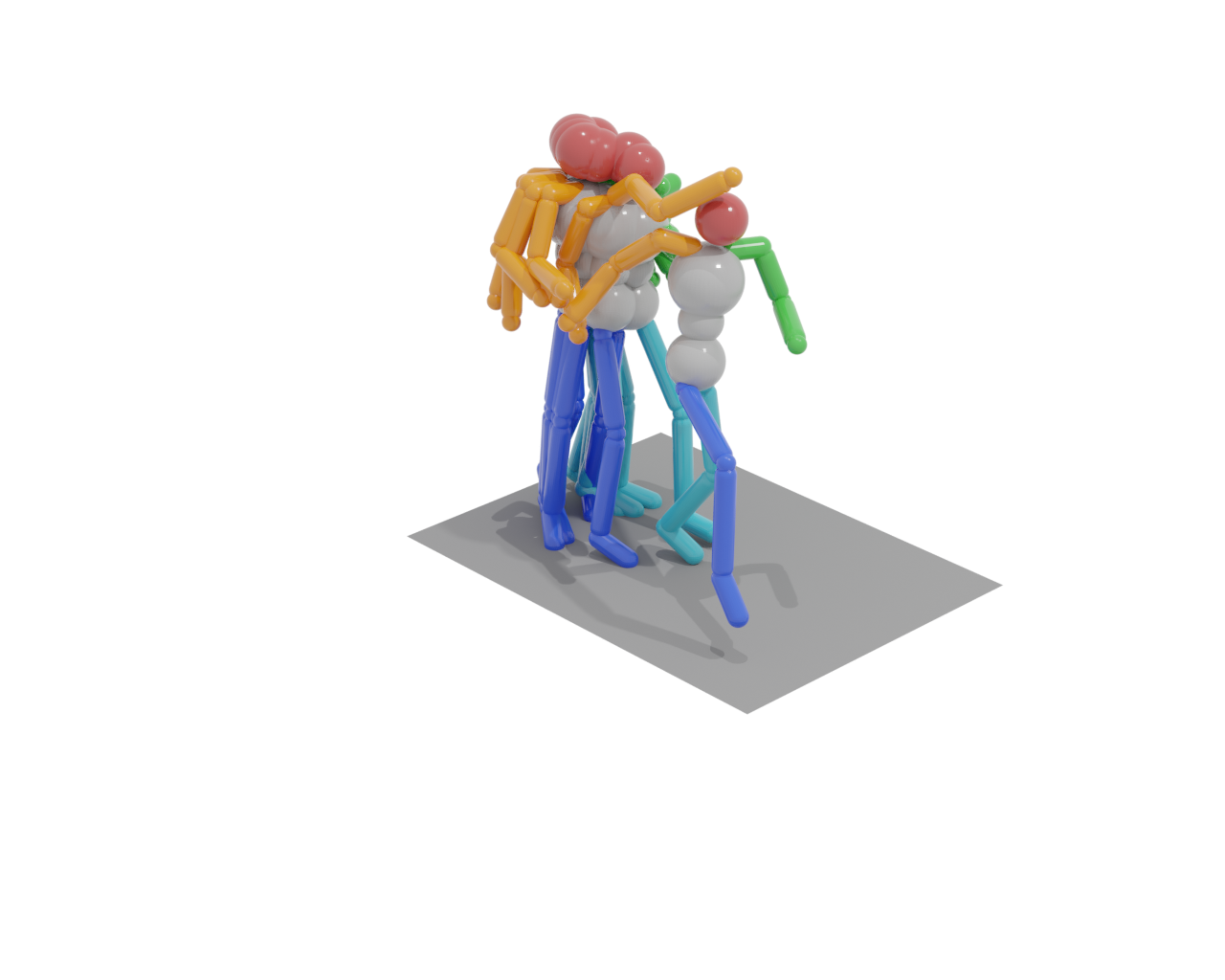}
&
\includegraphics[width=5.0cm,height=3.2cm,keepaspectratio]{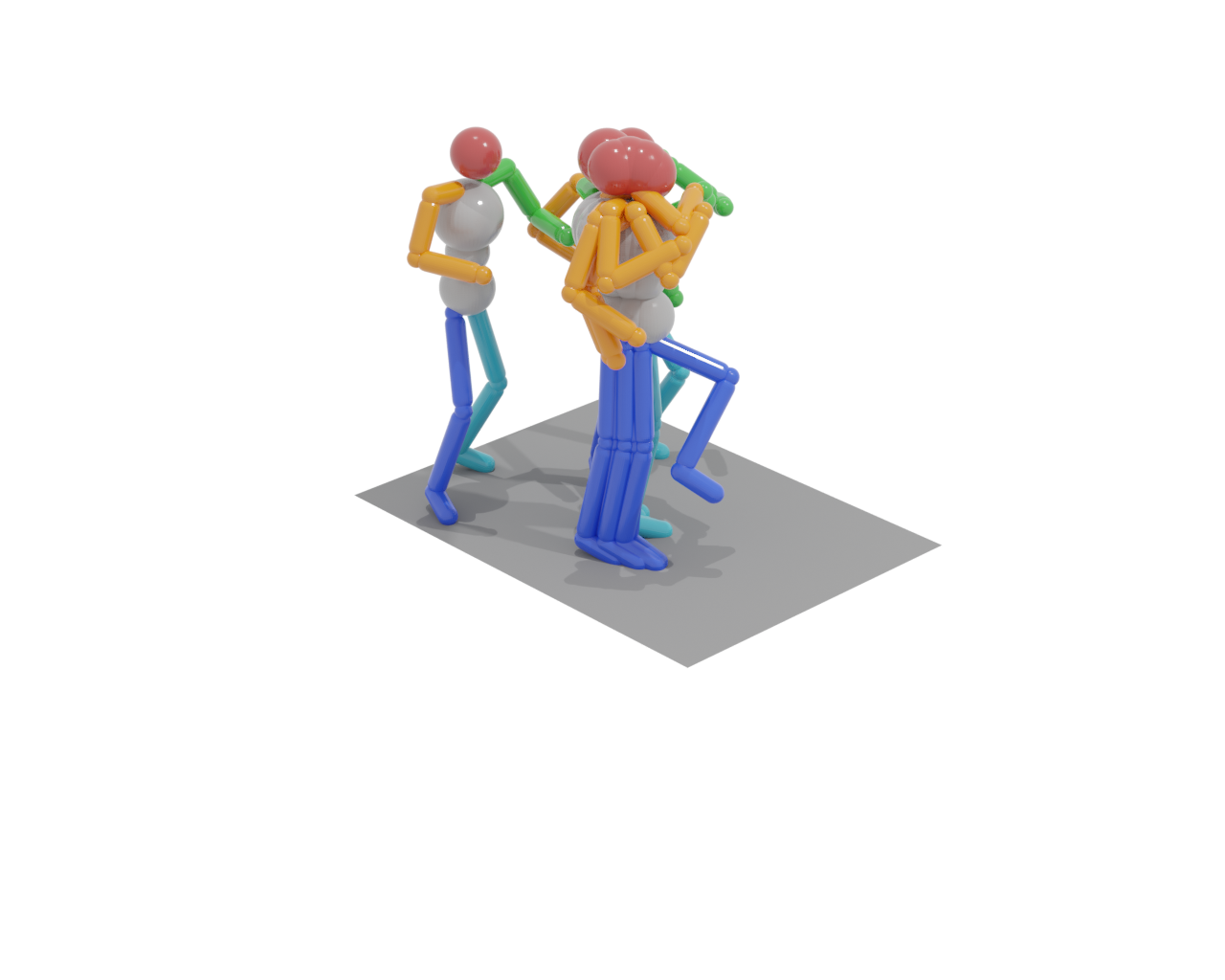}
\\

\bottomrule
\end{tabular}

\caption{Qualitative comparison (Part II).}
\label{tab:safemo_qualitative_simple_2}
\vspace{-0.8em}
\end{table*}

\section{More Qualitative Results}

We present qualitative comparisons between Motion Agent and our method MoRL in Table \ref{tab:safemo_qualitative_simple_1} and Table \ref{tab:safemo_qualitative_simple_2}, where the evaluated prompts cover sequential actions, continuous trajectory following, long-horizon repetition, and complex full-body coordination. These scenarios are intentionally selected to assess a model’s ability to preserve semantic structure, temporal coherence, and spatial constraints over extended motion sequences.

Across both tables, Motion Agent can generate visually plausible motions for simple and weakly constrained prompts. However, when the textual descriptions require explicit temporal ordering, global path planning, or fine-grained semantic modifiers, several systematic limitations become evident.

A prominent issue is Motion Agent’s difficulty in executing ordered multi-stage actions. For example, in the prompt “A person looks to the left then kicks something with their right foot” (Table \ref{tab:safemo_qualitative_simple_1}), Motion Agent tends to blur the two stages into a single ambiguous motion, where the head orientation change and the kicking action are not clearly separated in time. In contrast, MoRL produces a distinct head-turning phase followed by a well-timed right-foot kick, faithfully reflecting the sequential structure of the prompt.

Motion Agent also struggles to maintain global spatial trajectories over long horizons. In prompts such as “Walking slowly along the path shaped like an infinity symbol” and “A person walks along a curved path to the right” (Table \ref{tab:safemo_qualitative_simple_2}), Motion Agent frequently collapses the motion into locally plausible stepping patterns that fail to realize the intended global path, often resulting in clustered poses or near-stationary behavior. This indicates that the model prioritizes short-term kinematic validity over long-range spatial constraints specified in the text.

Another recurring limitation appears in long-horizon compositional execution. For instance, in the prompt “A person backflips three times in a row” (Table \ref{tab:safemo_qualitative_simple_2}), Motion Agent often produces incomplete or inconsistent repetitions, with noticeable degradation in motion amplitude and temporal rhythm across flips. This suggests difficulty in tracking and executing repeated action counts over extended sequences.

Finally, Motion Agent exhibits limited sensitivity to fine-grained motion modifiers.
In prompts such as “A person walks forward, slightly shifting to the right” and “A person walks forward with a side-to-side sway” (Table \ref{tab:safemo_qualitative_simple_1}), Motion Agent tends to default to a generic forward walking pattern, partially ignoring the subtle directional or stylistic constraints. Similarly, in “A person is practicing karate moves across the floor” (Table \ref{tab:safemo_qualitative_simple_2}), the generated motion often simplifies into repetitive gestures, losing the structured coordination implied by the prompt.

In contrast, MoRL consistently generates motion sequences that remain semantically faithful across all stages of the prompt. By explicitly modeling motion generation as a reasoning process, MoRL is able to plan long-term trajectories, preserve action ordering, and integrate fine-grained semantic constraints into motion execution. These qualitative results, observed consistently across Table \ref{tab:safemo_qualitative_simple_1} and Table \ref{tab:safemo_qualitative_simple_2}, demonstrate that MoRL better handles compositional, long-horizon, and structurally constrained motion generation, where Motion Agent exhibits inherent limitations.

\section{Ethical considerations}
\subsection{}
The datasets used in this work (HumanML3D, KIT-ML, and MotionHubV2) consist of motion capture data and textual descriptions of everyday human actions.
They do not contain personal identifiers such as names, addresses, or biometric identity information.
All textual annotations are action-level descriptions and do not include offensive, hateful, or sensitive personal content.
No additional personal data were collected as part of this work.

\subsection{}
\label{app:human_subjects}

This work includes a user study to evaluate the perceptual quality and semantic alignment of generated human motion sequences. 
Participants were asked to compare motions generated by different methods under the same textual description and provide subjective judgments based on predefined evaluation criteria.

Participants were provided with written instructions describing the evaluation task and criteria. 
They were instructed to focus on motion naturalness, semantic correctness with respect to the given text, and temporal coherence across the motion sequence. 
No deceptive instructions or sensitive content were involved, and participants were informed that they could stop the evaluation at any time.

Participants were recruited on a voluntary basis. 
The user study targeted adult participants and did not involve any demographic-based filtering. 
Participation was  compensated with a small reward consistent with standard academic user studies, and the compensation was considered reasonable given the short duration and low burden of the task.

Before participating, users were informed of the purpose of the study and that their responses would be used solely for academic research purposes. 
Only anonymized preference scores were recorded.  No personally identifiable information was collected, stored, or processed at any stage of the study.

The user study involved minimal risk and did not collect personal or sensitive data. 
Following common practice in prior human motion generation research, the study did not require formal institutional review board (IRB) approval.

\end{document}